\documentclass{article}

\usepackage[nonatbib, preprint]{neurips_2023}
\usepackage[square, numbers, sort&compress]{natbib}

\usepackage[utf8]{inputenc} 
\usepackage[T1]{fontenc}    
\usepackage{hyperref}       
\usepackage{url}            
\usepackage{booktabs}       
\usepackage{amsfonts}       
\usepackage{amssymb}
\usepackage[cmex10]{amsmath}
\usepackage{nicefrac}       
\usepackage{microtype}      
\usepackage{xcolor}         
\usepackage{graphicx} 
\usepackage{hyperref}
\usepackage{cleveref}
\usepackage{xr-hyper}
\usepackage{float}
\usepackage{balance}
\usepackage{mathtools}
\usepackage{bbm}
\usepackage{bm}
\usepackage{xcolor}
\usepackage{multirow}
\usepackage{plainsty}

\def\w{\bm{w}} 
\bmdefine\x{x} 
\bmdefine\X{X} 
\bmdefine\y{y} 
\def\1{\mathbbm{1}} 
\newcommand{\define}{:=}

\renewcommand{\hat}{\widehat}
\renewcommand{\tilde}{\widetilde}

\usepackage[ruled,vlined,algo2e]{algorithm2e}
\SetKwInput{Input}{Input}
\SetKwProg{kwSystem}{System executes}{:}{}
\SetKwProg{kwServer}{ConstUpdate}{:}{}
\SetKwProg{kwClient}{ClientUpdate}{:}{}

\def\mfg{\mbox{FedGFT}\xspace}
\usepackage{amsthm}
\theoremstyle{plain}
\newtheorem{theorem}{Theorem}[section]

\newtheorem{corollary}[theorem]{Corollary}
\theoremstyle{definition}
\newtheorem{definition}[theorem]{Definition}
\newtheorem{assumption}[theorem]{Assumption}
\theoremstyle{remark}
\newtheorem{remark}[theorem]{Remark}

\renewcommand{\citet}{\citep} 

\title{Mitigating Group Bias in Federated Learning: \\ Beyond Local Fairness}

\author{%
  Ganghua Wang \\
  School of Statistics\\
  University of Minnesota\\
  Minneapolis, MN 55455 \\
  \texttt{wang9019@umn.edu} \\
  \And
  Ali Payani \\
  Cisco Systems Inc. \\
  San Jose, CA, 95134  \\
  \texttt{apayani@cisco.com} \\
  \AND
  Myungjin Lee \\
  Cisco Systems Inc. \\
  San Jose, CA, 95134  \\
  \texttt{myungjle@cisco.com} \\
  \And
  Ramana Kompella \\
  Cisco Systems Inc. \\
  San Jose, CA, 95134  \\
  \texttt{rkompell@cisco.com} \\
}

\begin{document}

\maketitle

\begin{abstract}
The issue of group fairness in machine learning models, where certain sub-populations or groups are favored over others, has been recognized for some time. While many mitigation strategies have been proposed in centralized learning, many of these methods are not directly applicable in federated learning, where data is privately stored on multiple clients. To address this, many proposals try to mitigate bias at the level of clients before aggregation, which we call locally fair training. However, the effectiveness of these approaches is not well understood. In this work, we investigate the theoretical foundation of locally fair training by studying the relationship between global model fairness and local model fairness. Additionally, we prove that for a broad class of fairness metrics, the global model's fairness can be obtained using only summary statistics from local clients. Based on that, we propose a globally fair training algorithm that directly minimizes the penalized empirical loss. Real-data experiments demonstrate the promising performance of our proposed approach for enhancing fairness while retaining high accuracy compared to locally fair training methods.  
\end{abstract}

\section{Introduction} \label{sec:intro}
\hyphenation{through-out}

    As edge devices such as mobile phones and wearable devices have been heavily involved in our daily life, leveraging the enormous data collected by those devices and their computational resources to train machine learning models has attracted increasing research interest. 
    One challenge is that datasets collected by different devices are often forbidden to be shared due to communication costs and privacy concerns. Thus, classical centralized learning, where data is gathered and stored in a central database, is not suitable. 
    To address those challenges, federated learning~\citep{mcmahan2017communication, konevcny2016federated} has been proposed to train models in a decentralized manner.
    In federated learning, a global model is distributed to multiple clients, or edge devices, which update the model using their own data and send the updated model back to a central server. The server then aggregates the updated models to obtain a new global model and the process is repeated. 
    
    While significant progress has been made in the theory and application of federated learning~\citep{li2020federated}, most research has focused on improving the prediction accuracy of the global model.
    As these models are increasingly being used in areas that have a direct impact on people's lives, such as healthcare, finance, and criminal justice~\citep{berk2019accuracy, barocas2016big, becker2010economics}, the ethical implications of these models have attracted a lot of attention. In particular, it is crucial for the learned model to treat different groups in the population equitably.
    Nevertheless, it has been recognized that without careful consideration of group fairness \footnote{We use the words `fairness' and `bias' interchangeably throughout the paper. Increasing fairness means decreasing the bias.} the learned model may be biased~\citep{guion1966employment, cleary1968test, hutchinson201950} . 
    For example, the COMPAS algorithm~\citep{angwin2016machine}, which assigns recidivism risk scores to defendants based on their criminal history and demographic attributes, was found to have a significantly higher false positive rate for black defendants than white defendants, thereby violating the principle of equity on the basis of race. This highlights the risk of similar issues to arise in other applications such as university admissions and job screenings, which can negatively impact diversity and ultimately harm the society.

    Though bias mitigation has been extensively studied in the centralized setting
    ~\citep{kamiran2012data, kamishima2012fairness, zafar2017fairness, zhang2018mitigating, hardt2016equality, lohia2019bias, caton2020fairness}, it remains under-explored for federated learning.
    Many of the currently proposed algorithms try to reduce the global model's bias by minimizing the bias of local models~\citep{abay2020mitigating, ezzeldin2021fairfed}, hereinafter referred to as \textit{locally fair training} (LFT). Because the global model is the average of local models, LFT hopes the global model is fair as long as local models are fair. 
    However, theoretical understanding of LFT is limited, such as under what conditions LFT is effective. One main challenge of analysis is obtaining the fairness measure for the global model without sharing original data across devices. In this work, we tackle this challenge for a particular class of fairness metrics. Based on that, we show LFT works well for near-homogeneous clients. Furthermore, we propose a globally fair training algorithm that directly maximizes the global model's fairness. 
    
    Our contributions are three-fold as summarized below. 
    \begin{enumerate} 
        \item We study the relationship between fairness of local and global models, for the first time revealing their underlying theoretical connection. In general, global fairness and local fairness do not imply each other.
        Nevertheless, for proper group-based fairness defined in Section~\ref{sec:result}, the global fairness value is controlled by the local fairness values and the data heterogeneity level. This result explains the success of LFT methods in the setting of near-homogeneous clients for common fairness metrics, such as demographic parity and equal opportunity.
        \item We formulate the definitions of group-based and proper group-based fairness metrics. For proper group-based metrics, the global fairness value can be expressed as a function of fairness-related statistics calculated by local clients solely. 
        This property enables us to calculate the global fairness value without directly accessing local datasets.
        In particular, those fairness-related statistics are not local fairness values, distinct from all existing works. 
        \item We propose a globally fair training method named \mfg for proper group-based metrics. \mfg goes beyond LFT by directly solving a regularized objective function consisting of the empirical prediction loss and a penalty term for fairness. Additionally, it applies to clients with arbitrary data heterogeneity. Numerical experiments on multiple datasets show that \mfg significantly reduces the bias of the global model while retaining high prediction accuracy.
    \end{enumerate}

\section{Preliminaries} \label{sec:literature}
 \subsection{Federated learning} \label{subsec:fl}
        There is a large body of literature on federated learning~\citep{li2020federatedl} since proposed by \citep{konevcny2016federated,mcmahan2017communication}. It aims to train a global machine learning model while keeping the training data privately on edge devices, also named local clients.
        Suppose there are $K$ clients in total, and the $k$-th client owns $n_k$ training data $\bigl\{\X_k^{(i)}, Y_k^{(i)} \bigr\}_{i=1}^{n_k}$, where $\X$ is the predictor and $Y$ is the response.
        Let $l(\cdot, \cdot)$ be a loss function, federated learning aims to solve the following empirical risk minimization problem:
        \begin{align*} 
            &\min_{\theta} \sum_{k=1}^K \frac{n_k}{n} L_k(\theta),  \numberthis \label{eq:global_fl}
            \text{ where } L_k(\theta) = \frac{1}{n_k} \sum_{i=1}^{n_k} l(f(\X_k^{(i)};\theta), Y_k^{(i)}), n=\sum_{k=1}^K n_k.
        \end{align*}
        Here, $L_k(\theta)$ is the empirical risk of the $k$-th client, $f(\cdot;\theta)$ is a parameterized model.
        
        The original idea of federated learning is training the model on each client using its local dataset for several updating steps, then aggregating the local models on the central server to obtain a global model, and repeating the above procedure until meeting the terminating conditions. More specifically, at each communication round $t$, the server first propagates the parameters $\theta^t$ of the current global model 
        to the clients. Then, each client will perform $E$ epochs of local updates to get $\theta^{t,E}_k, k=1,\dots, K$. Finally, the server will aggregate $\theta^{t,E}_k$'s to a new global model with parameter $\theta^{t+1}$.

    \subsection{Group fairness}
            There are many different interpretations of fairness~\citep{caton2020fairness, mehrabi2021survey}. In this paper, we focus on group fairness, which ensures that the model will not have discriminatory behavior towards certain groups. 
            For simplicity, we consider a binary classification task with the outcome $Y \in \{0,1\}$, and sensitive group $A \in \{0,1\}$.
            There are two major categories of group fairness quantification~\citep{corbett2018measure}. The first category is based on the classification parity, which means a measure of the prediction error is equal across different groups. 
            For example, statistical parity~\citep{kamishima2012fairness, feldman2015certifying}, also known as demographic parity, requires that the distribution of the prediction $\hat{Y}$ conditional on the sensitive group is the same. 
            In other words, $\P(\hat{Y}=1|A=0)=\P(\hat{Y}=1|A=1)$. 
            Another example is equal opportunity~\citep{hardt2016equality}, which requires the same true positive rate across groups, i.e., $\P(\hat{Y}=1|A=0,Y=1)=\P(\hat{Y}=1|A=1,Y=1)$. The second category is calibration~\citep{pleiss2017fairness}. 
            A model is well-calibrated or achieves test fairness if the true outcome is independent of the group given the predicted value. 
            We note that different fairness definitions may be incompatible; actually, it is impossible to achieve multiple fairness goals simultaneously~\citep{kleinberg2016inherent}. 


\section{Problem formulation} \label{sec:form}
       This paper considers a binary classification task with outcome $Y \in \{0,1\}$. Suppose the predictors $\X = (X_1, \dots, X_p)^\T \in \real^p$ are $p$-dimensional variables. Without loss of generality, we assume the first predictor to be the sensitive attribute as $A = X_1\in\{0,1\}$, and other predictors are non-sensitive. 
    We consider a heterogeneous scenario that there are $K$ clients, and the $k$-th client's training data $\bigl\{\X_k^{(i)}, Y_k^{(i)} \bigr\}_{i=1}^{n_k}$ is IID generated from a distribution $\mathcal{D}_k$. 
    The empirical distribution of $\bigl\{\X_k^{(i)}, Y_k^{(i)} \bigr\}_{i=1}^{n_k}$ is denoted as $\hat{\mathcal{D}}_k$. 
    Our goal is to learn a function $f: \real^p \to [0, 1]$ from data, where $f(\X)$ is regarded as the predicted probability of $\P(Y=1\mid X)$. 
    The accuracy of the learned function $f$ is evaluated by the prediction risk $\E \{l(f(\X), Y)\}$, where $\E$ denotes expectation, and $l(\cdot, \cdot)$ is a loss function, such as the cross entropy loss. 
    As for the fairness measure, we define the following group-based fairness metrics. 
    \begin{definition}[Group-based fairness metrics]\label{def:fair}
        $F(f, \mathcal{D})$ is a group-based fairness metric if it is in the form of 
        \begin{align*}
            F(f, \mathcal{D}) = \biggl| \frac{a(f, \mathcal{D})}{b(f, \mathcal{D})}- \frac{c(f, \mathcal{D})}{d(f, \mathcal{D})} \biggr|,
        \end{align*}
        where $a(f, \mathcal{D})$ and $b(f, \mathcal{D})$ are some expectations on the event $\{A=0\}$,
        $c(f, \mathcal{D}), d(f, \mathcal{D})$ are some expectations on the event $\{A=1\}$. Moreover, we have the range of $a,b,c,d$, $a/b$ and $c/d$ be $[0,1]$, where $a,b,c,d$ stands for four functions omitting the arguments. 
    \end{definition}
        Clearly, a smaller $F(f, \mathcal{D})$ indicates higher model fairness and smaller model bias.
        The concept of group-based fairness metrics~\citep{caton2020fairness} is motivated by the observation that many fairness metrics are the disparity between group-specific quantities, such as the confusion-matrix based probabilities~\citep{kim2020fact}. But it is the first time a theoretical formulation is given to group-based fairness metrics.   
        By Bayes' theorem, those group-specific quantities can be further written as the ratio of two expectations.
     Definition~\ref{def:fair} includes many common measures, such as the following three. We can verify this by checking \Autoref{tab:abcd}, with full details in supplementary document.
    
    \textbf{Statistical Parity (SP).} It is defined as $F(f, \mathcal{D}) = \abs{\P(\hat{Y}=1|A=0)-\P(\hat{Y}=1|A=1)}$. 
    \begin{table}
        \small
        \centering
        \caption{The associated bi-linear functions of three fairness metrics.}
        \label{tab:abcd}
        \resizebox{\textwidth}{!}{
        \begin{tabular}{lccccc}
            \toprule
             \textbf{Metrics} & $a(f, \mathcal{D})$ & $b(f, \mathcal{D}) $ & $c(f, \mathcal{D})$ & $d(f, \mathcal{D}) $ \\
             \midrule
             Statistical Parity  & $\P(\hat{Y}=1, A=0)$ &  $ \P(A=0)$ & $\P(\hat{Y}=1, A=1)$ & $ \P(A=1)$ \\ \midrule
             Equal opportunity & $\P(\hat{Y}=1, Y=1, A=0)$& $\P(Y=1, A=0)$&$\P(\hat{Y}=1, Y=1, A=1)$&$\P(Y=1, A=1)$\\ \midrule
             Well-Calibration  &$ \P(Y=1, \hat{Y}=1, A=0)$&$\P(\hat{Y}=1, A=0)$&$\P(Y=1, \hat{Y}=1, A=1)$&$\P(\hat{Y}=1, A=1)$\\ \bottomrule
        \end{tabular}
        }
    \end{table}

    \textbf{Equal Opportunity (EOP).} $F(f, \mathcal{D}) = \abs{\P(\hat{Y}=1|A=0,Y=1)-\P(\hat{Y}=1|A=1,Y=1)}$. 
    
    \textbf{Well-Calibration.} $F(f, \mathcal{D}) = \abs{\P(Y=1|A=0,\hat{Y}=1)-\P(Y=1|A=1,\hat{Y}=1)}$. 


    In practice, since the true underlying distribution is typically unknown, we take the empirical estimation $F(f, \hat{\mathcal{D}}_k)$ as a surrogate for the local fairness of the $k$-th client,
    and use $F(f, \hat{\mathcal{D}})$ as the global fairness, where $ \hat{\mathcal{D}} = \sum_{i=1}^K w_k \hat{\mathcal{D}}_k, w_k =n_k /n,  \ n = \sum_{i=1}^k n_k.$

    Recall that the learned function $f$ is expected to be both accurate (with respect to the classification task) and fair (with respect to the sensitive group $A$). 
    Locally fair training is one approach to extend bias mitigation methods from the centralized setting to the federated learning setting. 
    Essentially, it minimizes the bias of each local client at each communication round and expects that the aggregation of locally fair models will yield a globally fair model. To better understand the effectiveness of locally fair training methods, we are going to study the following two fundamental questions in next sections:
    \begin{enumerate}
        \item What is the relationship between fairness of local models and the global model? 
        \item Is there an algorithm that directly targets improving global fairness?
    \end{enumerate}
    The answer to the first question is local fairness does not imply global fairness in general. Nevertheless, for a proper group-based fairness metric, which is defined in Section 4, we show that global fairness can be controlled by local fairness and data heterogeneity. To our best knowledge, this is the first work to systematically study the relationship between local and global fairness. As for the second question, we propose such an algorithm called \mfg in Section 5. 
    
    

\section{Locally fair training} \label{sec:result}
\def\DH{\textsc{DH}}
    This section explores the relationship between local and global fairness, which helps us in analyzing the locally fair training methods. 
    The idea of minimizing the biases of local models is appealing at the first glance, based on an intuition that global fairness will be guaranteed if all local models are fair. \citep{chu2021fedfair} proved that this intuition is true for homogeneous clients and a special fairness metric, accuracy disparity. However, we show that it does not hold in general.

    \begin{theorem}[In general, Global $\neq$ local]\label{thm:general}
        Suppose $F$ is a group-based fairness metric. For any $0\leq C \leq 1$, there exist a model $f$ and local data distributions $\{\hat{\mathcal{D}}_k, k=1,\dots,K\}$ such that $F(f, \hat{\mathcal{D}}_k)=0$ for all $k$, and $F(f, \hat{\mathcal{D}}) \geq C$.
        Conversely, for any $0\leq C \leq 1$, there also exists another set of $f$ and $\{\hat{\mathcal{D}}_k, k=1,\dots,K\}$ such that $F(f, \hat{\mathcal{D}}) = 0$, and $F(f, \hat{\mathcal{D}}_k) \geq C$ for all $k$.
    \end{theorem}
    \begin{corollary}\label{coro:general}
        Suppose $F$ is a group-based fairness metric, then $F(f, \hat{\mathcal{D}})$ cannot be written as a linear combination of $\{F(f, \hat{\mathcal{D}}_k), k=1,\dots,K\}$.
    \end{corollary}
    
    All proofs are included in the supplementary document. \Autoref{thm:general} means that locally fair models do not imply a fair global model and vice versa. Therefore, locally fair training methods are not always effective in general. Additionally, Corollary~\ref{coro:general} indicates that global fairness cannot be obtained from a simple average of local fairness.
    Simpson's paradox~\citep{blyth1972simpson, bickel1975sex} is an excellent example to illustrate that the local property cannot represent that of the global, as shown in \Autoref{tab:simpson}. Suppose a college has two departments, A and B, which accept applications from high school students. 
    Here, gender is the sensitive group, and each department is considered a client. Although the acceptance rate is the same between males and females (i.e., SP is zero) for both departments, the overall acceptance rate is significantly biased toward males.
    \begin{table}[H]
        \scriptsize
        \centering
        \caption{The admission example of gender bias.} \label{tab:simpson}
        \begin{tabular}{lrrrr}
            \toprule
            \multirow{2}{*}{Department} & \multicolumn{2}{c}{Female} & \multicolumn{2}{c}{Male}\\
            \cmidrule(l){2-3} \cmidrule(l){4-5}
            &Applicants & Acceptance & Applicants & Acceptance \\
            \midrule
            A & 90 & 20\% & 10  & 20\% \\
            \midrule
            B & 10 & 80\% & 90  & 80\% \\
            \midrule
            Total & 100 & 26\% &100& 74\% \\
            \bottomrule
        \end{tabular}
    \end{table}

     The major factor that may lead to the failure of LFT is data heterogeneity, as
    revealed in the following two theorems.
    \begin{theorem}\label{thm:perfect_fairness}
        Suppose $F$ is a group-based fairness metric, $f$ is non-degenerated, and $\mathcal{D} = \sum_{i=1}^K w_k \mathcal{D}_k$, where $w_k$ is the aggregation weight. 
        A necessary and sufficient condition for $F(f, \mathcal{D})=0$ holds for any $w_k$ is that there exists a constant $C$ such that $a_k/b_k = c_k/d_k = C$ for all $k$, 
        where $g_k = g(f, \mathcal{D}_k)$ for $ g \in \{a,b,c,d\}$.
    \end{theorem}
    \Autoref{thm:perfect_fairness} implies that if the global model is perfectly unbiased regardless of sample sizes of clients, then all local models are also unbiased.
    Note that this is assured if the data distributions of different clients are homogeneous. 
    Inspired by \Autoref{thm:perfect_fairness}, a natural idea to evaluate data heterogeneity is the maximum difference of $a_k/b_k$ (also $c_k/d_k$) among all clients. However, those quantities involve the global model $f$, which is unknown before the training. Thus, we introduce the following concepts to decouple with $f$. 
    \begin{definition}[Proper Group-based fairness metrics]\label{def:proper}
         A group-based fairness metric $F(f, \mathcal{D})$ is proper if the corresponding $b(f, \mathcal{D})$ and $d(f, \mathcal{D})$ are degenerated with respect to $f$. In other words, there exist a function $b'$ such that $b(f, \mathcal{D})= b'(\mathcal{D})$, and similarly for $d(f, \mathcal{D})$.
    \end{definition}
    \begin{definition}[Data heterogeneity with respect to $F$]
        For a proper group-based fairness metric $F$, let $b = \sum_k w_k b_k$ and $d = \sum_k w_k d_k.$
        The data heterogeneity coefficient is defined as 
        \begin{align*}
            \DH(\{\hat{\mathcal{D}}_k, k=1,\dots, K\}) = \max_{k} \biggl|\frac{d}{b}\frac{b_k}{d_k}-1\biggr|.
        \end{align*}
    \end{definition}
    Many fairness measures are proper such as SP and EOP, while calibration is not proper, as indicated by \Autoref{tab:abcd}. 
    For proper metrics, $\DH$ measures the relative variation of two data-determined statistics $b_k$ and $d_k$, hence reflects the influence of data heterogeneity.
    More importantly, $\DH$ relates the global and local fairness as follows.
    \begin{theorem}[Near IID, local implies global]\label{thm:neariid}
        Suppose $F$ is proper, the data heterogeneity coefficient of clients' data is $\beta$,
        and $F(f, \hat{\mathcal{D}}_k) \leq \alpha$ for all $k$, then $F(f, \hat{\mathcal{D}}) \leq \alpha + \beta. $
    \end{theorem}
    \Autoref{thm:neariid} shows that the global fairness is upper-bounded by the local fairness and data heterogeneity level for proper group-based fairness metrics. This upper bound is tight when data heterogeneity level $\beta$ is small. On the one hand, it justifies the success of locally fair training methods in the region of near-homogeneous situations; on the other hand, it implies that locally fair training may fail when data distributions are highly different. Thus, together with \Autoref{thm:general}, we provide a fundamental understanding of the first question asked in \Autoref{sec:form}. Furthermore, the proper group-based fairness metrics provide the possibility to calculate the global fairness value using information from local clients. 
    In the next section, we will utilize this observation and propose a globally fair training algorithm, which answers the second question in \Autoref{sec:form}.
    
    \section{Beyond local fairness}\label{sec:global}
        Recall the ultimate goal is to obtain a fair and accurate global model. 
        In the centralized setting, it is standard to minimize the empirical loss with fairness regularization~\citep{bellamy2019ai, berk2017convex} as follows:
        \begin{align}
        \min_{\theta} L(\theta) \define \sum_{k=1}^K \frac{n_k}{n} L_k(\theta) + \lambda J(F(f(\cdot ;\theta); \hat{\mathcal{D}})), \label{eq:obj_pen}
        \end{align}
        where $J(\cdot)$ is a regularization function.
        Without the fairness regularization, \Autoref{eq:obj_pen} is reduced to Eq.~\ref{eq:global_fl}, where the gradient of the global objective function can be calculated or estimated by aggregating the gradients of local objective functions $L_k$. FedAvg~\citep{mcmahan2017communication} is inspired by this observation, which performs the gradient descent algorithm on each local client and then aggregates local models. 
        Thus, to generalize federated learning algorithms to fairness-regularized objective \Autoref{eq:obj_pen},
        the challenge is how to obtain the gradient of global fairness using summary statistics from local clients. However, as we showed in Corollary~\ref{coro:general}, the global fairness value cannot be simply represented by local fairness in general. 

        Fortunately, the next theorem shows that for a proper group-based fairness metric $F$, the gradient of \Autoref{eq:obj_pen} can be calculated from the gradients of fairness-specific local objectives.
        \begin{theorem}\label{thm:gradient}
            Let $b = \sum_k w_kb_k$, $d = w_kd_k$, and $F_k = a_k/b - c_k/d$.
            For a proper group-based fairness metric $F$, we have
            \begin{align*}
                 \tilde{F} = \sum_{k=1}^K w_k F_k, \quad F(f,\hat{\mathcal{D}}) = \abs{\tilde{F}}, \quad
                 \nabla_{\theta} J(F(f,\hat{\mathcal{D}}))=C_{\theta} \sum_{k=1}^K w_k \nabla_{\theta} F_k, 
            \end{align*}
            where $C_{\theta}=\sign(\tilde{F}) \nabla_F J(F(f,\hat{\mathcal{D}})) $ is a constant of $F_k$'s.
        \end{theorem}
        \Autoref{thm:gradient} indicates that we can apply the gradient descent algorithm to minimize \Autoref{eq:obj_pen}, similar to the centralized setting. 
        Specifically, at each round $t$, the local client should optimize the following fairness-augmented objective
        \begin{align*}
           \min_{\theta} L_k(\theta)+\lambda C_{\theta^{t-1}} F_k(\theta),
           \numberthis \label{eq:global_update}
        \end{align*}
        then the aggregation of local models will give the correct gradient descent update of the global objective function. 

        Motivated by \Autoref{thm:gradient},
        we propose a globally fair training method named \mfg and summarize it in Algorithm~\ref{alg:m}. We note that \mfg can incorporate most existing FL algorithms.
        While the aggregation method on the server side and the optimization tool on the client side remain the same, \mfg adapts the local objective function to the fairness regularization. Moreover, \mfg also applies to the situation where clients are purely from one group (for example a client with all points from Group A, and another client with all points from group B), which is not allowed for LFT.
        \begin{remark}
        Many fairness metrics are not differentiable. Taking SP for example, $a_k=\P(\hat{Y}=1, A=1) = \sum_{i=1}^{n_k} \1_{f(\X_k^{(i)})>0.5}\1_{{A_k}^{(i)}=0}$ is not differentiable. A common strategy is using a surrogate, such as the softmax score $\sum_{i=1}^{n_k} f(\X_k^{(i)})\1_{{A_k}^{(i)}=0}$. 
        \end{remark}
        
        \begin{algorithm}[tb]
        \SetAlgoLined
\DontPrintSemicolon
        \caption{(\mfg) Federated learning with globally fair training}\label{alg:m}
            \Input{Communication rounds $T$, learning rate $\eta$, local training epochs $E$, batch size $B$, penalty parameter $\lambda$.}

            \kwSystem{}{
            Initialize the global model parameters $\theta^0$
            
            
    \For{\textup{each communication round }$t = 1,2,\dots T$}{
        Sample a subset $S_t \subseteq \{1,\dots, K\}$

        Update the constant $C_{\theta^{t-1}} \leftarrow$\textbf{ConstUpdate}$(\theta^{t-1})$
        
        \For{\textup{each client }$k \in S_t$\textup{ \textbf{in parallel}}}{
    
            Receive the model parameters $\theta_k^{t,0} = \theta^{t-1}$ from the server
    
            $\theta_k^{t,E} \leftarrow$ \textbf{ClientUpdate}$(\theta_k^{t,0}, C_{\theta^{t-1}})$
        }
        6
        Server update global model $\theta^t$ by aggregating $\theta_k^{t,E}, k \in S_t$, using any FL algorithm

    }
    Return the final global model $f(\cdot;\theta^T)$
}
\kwServer{\textup{$(\theta^t)$}}{
\For{\textup{each client }$k\in S_t$\textup{ \textbf{in parallel}}}{
Calculate $F_k(\theta^t)$ 
}
$\tilde{F} \leftarrow \sum_{k\in S_t} w_k F_k$

Return $\nabla_F J(\abs{\tilde{F}})\sign(\tilde{F})$
}
\kwClient{\textup{$(\theta_k^{t,0}, C_{\theta^{t-1}})$}}{
    \For{\textup{each local epoch $e$ from 1 to $E$}}{
    %
    \For{\textup{each batch $b$ from 1 to $B$}}{
    Update model parameters by any FL algorithm with local objective Eq.~\ref{eq:global_update}
    }
    }
    Return $\theta_k^{t,E}$ 
}
        \end{algorithm}

        Furthermore, we prove that \mfg will converge to a stationary point when we use gradient-based optimization tools. The complete statement and proof are included in the supplementary document.
        \begin{theorem}[Covergence analysis] \label{thm:converge}
        Suppose the local clients apply one-step stochastic gradient descent to optimize Eq.~\ref{eq:global_update}, and the global server updates the global model by averaging a random subset of local models. Let $\theta^t$ be the parameter of the global model at round $t$. 
                Under mild assumptions, for a step-size sequence $\{\eta_t, t=0,\dots,T-1 \}$, we have 
    \begin{align*}
        \min_{t=0,\dots,T-1} \E ( \norm{\nabla L(\theta^t)}^2 ) \leq C\biggl(\frac{1}{\sum_{t=0}^{T-1} \eta_t} + \frac{ \sum_{t=0}^{T-1}\eta_t^2}{\sum_{t=0}^{T-1} \eta_t}\biggr),
    \end{align*}
    where $C$ is a constant independent of $T$ and $\{\eta_t, t=0,\dots,T-1\}$. 
        \end{theorem}
        \begin{corollary}\label{coro:covergence}
        The choice of $\eta_t=O(1/t), t\geq 1$ yields 
    $ 
         \min_{t=0,\dots,T-1} \E ( \norm{\nabla L(\theta^t)}^2 ) \leq O(1/\log(T)),
    $ where $O$ is the big-$O$ notation.
    The choice of $\eta_t = O(t^{-1/2})$ yields
    $ 
        \min_{t=0,\dots,T-1} \E ( \norm{\nabla L(\theta^t)}^2 ) \leq O(\log(T)T^{-1/2}).
    $ 
    Furthermore, if the gradient descent algorithm is used for optimization instead of stochastic gradient descent, then choosing $\eta_t=\eta_0$ yields a faster rate:
    $ 
        \min_{t=0,\dots,T-1} \E ( \norm{\nabla L(\theta^t)}^2 ) \leq O(T^{-1}).
    $ 
\end{corollary}
    
\section{Experiments} \label{sec:exp}
    \def\local{LRW}

\subsection{Setup}
    For all the experiments below, we train a binary classification model using four methods: baseline method (`FedAvg', ~\citet{mcmahan2017communication}), state-of-art fairness-aware FL (`FairFed', ~\citet{ezzeldin2021fairfed}), locally reweighing (`\local'), and our proposed globally fair method (`$\mfg$'). 
    We consider three datasets, Adult~\citep{Dua:2019}, COMPAS~\citep{angwin2016machine}, and CelebA~\citep{liu2015faceattributes}. For each dataset, we split the training data as follows. First, we generate the proportion of each combination of the group variable $A$ and response variable $Y$ for each client from a Dirichlet distribution $\textrm{Dir}(\alpha)$. A larger $\alpha$ implies more homogeneous clients.  Then, we randomly assign the corresponding proportion of data points to each client. Throughout this section, we consider $\alpha=\{0.5, 5, 100\}$.
    The test criteria are test accuracy (using zero-one loss) and the global fairness metric. Note that we conduct the experiments using both SP and EOP as the fairness metric, respectively.
    All experiments are replicated $20$ times. 
    Further details of the training are included in the supplementary document. 

     \textbf{Adult dataset} The Adult dataset contains the income level and demographic attributes of $48842$ people. We train a logistic regression model to predict a binary response `Income' (high or low) with $14$ continuous and categorical predictors. 
        The predictor `Race' (white or non-white) is considered the sensitive group. 
        We choose local update epoch $E=1$, clients number $K=10$, communication rounds $T=20$. Note that each epoch will divide local datasets into several batches and thus perform multiple steps of local update.
        The hyper-parameter for `FairFed' is chosen from $\{0.1 ,1 , 10\}$ with cross-validation,
        and for `\mfg' is chosen from $\{1, 10, 20, 50\}$.

        \textbf{COMPAS dataset}
    This dataset includes ten demographic attributes of $6172$ criminal defendants and whether they recidivate in two years. A logistic model is trained to predict recidivism, and gender is the sensitive variable. All other settings are the same as the Adult experiment above.

    \textbf{CelebA dataset}
        This dataset contains $202,599$ face images, and each image has $40$ binary attributes. In this experiment, we train a ResNet18 model~\citep{he2016deep} targeting at classifying the `Smiling' attribute (yes or no), and take `Male' (yes or no) as the sensitive attribute. To speed up the training process, we randomly select $10,000$ images for training and $6,000$ for testing in each replicate. All 
        other settings are the same as the Adult dataset.

\vspace{-0.2cm}
\subsection{Results}
\newcommand{\tour}{$\mfg^\dag$}

        \begin{table}[tb]\scriptsize
            \centering
            \caption{The average accuracy and bias (standard error in parentheses) on three datasets, under three heterogeneity levels and two fairness metrics. The proposed method is marked by $\dag$.}
            \label{tab:adult}
\begin{tabular}{lllllllllll}
\toprule
    \multicolumn{2}{c}{Dataset} &  \multicolumn{3}{c}{Adult} & \multicolumn{3}{c}{COMPAS} & \multicolumn{3}{c}{CelebA} \\ \cmidrule(l){3-5} \cmidrule(l){6-8} \cmidrule(l){9-11} 
  $\alpha$    & Method  &   Acc ($\uparrow$) &         SP ($\downarrow$) &         EOP ($\downarrow$) &   Acc($\uparrow$) &         SP ($\downarrow$) &         EOP ($\downarrow$)  &   Acc ($\uparrow$) &         SP ($\downarrow$) &         EOP ($\downarrow$)    \\
\midrule
\multirow{4}{*}{$0.5$} & FedAvg  &     79.8 (1.3)  &    2.9 (1.6)  &    5.2 (2.6)  &    57.4 (8.4)  &    4.5 (3.4)  &    3.9 (2.8)  &      91.4 (0.7)  &    14.7 (3.2)  &    7.5 (2.7)  \\
                       & \local &     79.2 (2.1)  &    2.0 (1.2)  &    5.2 (1.9)  &    57.4 (6.1)  &    3.4 (2.2)  &    2.9 (2.5)  &      91.9 (0.4)  &    13.7 (1.0)  &    1.3 (0.7)  \\
                       & FairFed  &    67.9 (19.1)  &    1.6 (1.7)  &    1.8 (2.8)  &    57.4 (6.0)  &    3.5 (4.3)  &    9.0 (7.6)  &      91.7 (0.4)  &    13.8 (3.0)  &    7.7 (2.9)  \\
                       & \tour   &     80.0 (1.4)  &    0.8 (0.7)  &    0.9 (0.7)  &    56.7 (6.3)  &    1.0 (0.5)  &    1.3 (0.9)  &      88.9 (3.6)  &     8.1 (5.1)  &    1.7 (2.3)  \\ \midrule
\multirow{4}{*}{$5$}   & FedAvg  &     81.4 (0.6)  &    4.8 (0.4)  &    3.7 (0.5)  &    65.3 (2.9)  &    5.0 (2.4)  &    6.6 (1.9)  &      92.0 (0.3)  &    13.6 (0.4)  &    5.8 (0.4)  \\
                       & \local &     81.2 (0.7)  &    2.9 (0.3)  &    6.1 (0.5)  &    64.7 (2.4)  &    4.1 (2.0)  &    3.7 (1.3)  &      91.8 (0.3)  &    13.8 (0.2)  &    0.4 (0.2)  \\
                       & FairFed  &     78.8 (3.6)  &    2.8 (1.5)  &    4.8 (0.8)  &    64.3 (2.9)  &    3.9 (2.1)  &    3.1 (1.5)  &      91.9 (0.3)  &    13.8 (0.3)  &    6.0 (0.5)  \\
                       & \tour   &     80.9 (0.7)  &    0.7 (0.1)  &    0.2 (0.1)  &    64.5 (2.1)  &    0.9 (0.4)  &    1.1 (0.3)  &      90.7 (0.5)  &     4.9 (1.6)  &    0.7 (0.4)  \\ \midrule
\multirow{4}{*}{$100$} & FedAvg  &     81.4 (0.6)  &    4.7 (0.3)  &    3.3 (0.4)  &    65.9 (1.1)  &    8.2 (0.9)  &    7.8 (1.8)  &      92.0 (0.3)  &    13.6 (0.2)  &    5.7 (0.3)  \\
                       & \local &     81.1 (0.4)  &    2.9 (0.3)  &    6.5 (0.7)  &    66.4 (0.9)  &    5.8 (1.7)  &    5.0 (1.2)  &      91.9 (0.3)  &    13.8 (0.1)  &    0.2 (0.1)  \\
                       & FairFed &     81.6 (0.9)  &    2.8 (0.4)  &    5.5 (1.2)  &    65.7 (1.8)  &    4.9 (1.9)  &    4.6 (1.3)  &      91.4 (0.4)  &    13.7 (0.1)  &    5.8 (0.3)  \\
                       & \tour   &     81.0 (0.6)  &    0.6 (0.1)  &    0.2 (0.1)  &    65.2 (2.1)  &    1.3 (0.7)  &    1.1 (0.5)  &      90.8 (0.7)  &     6.4 (3.3)  &    0.3 (0.3)  \\
     
\bottomrule
\end{tabular}
        \end{table}

\begin{figure}[tb]
    \begin{minipage}[t]{.49\textwidth}
        \centering
        \includegraphics[width=\textwidth]{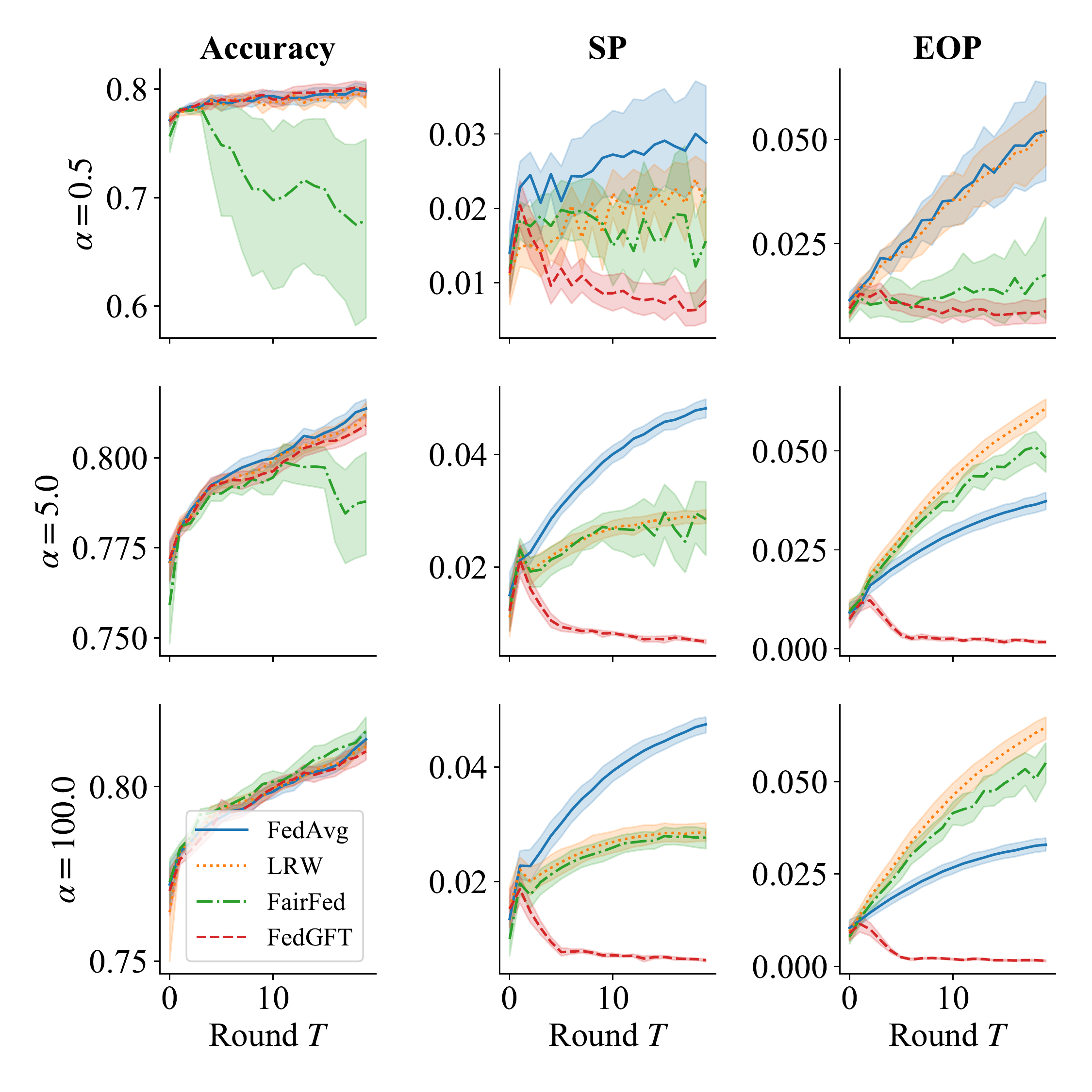}
        \caption{The accuracy and bias on the Adult dataset.}\label{fig:adult}
    \end{minipage}
    \hfill
    \begin{minipage}[t]{.49\textwidth}
        \centering
        \includegraphics[width=\textwidth]{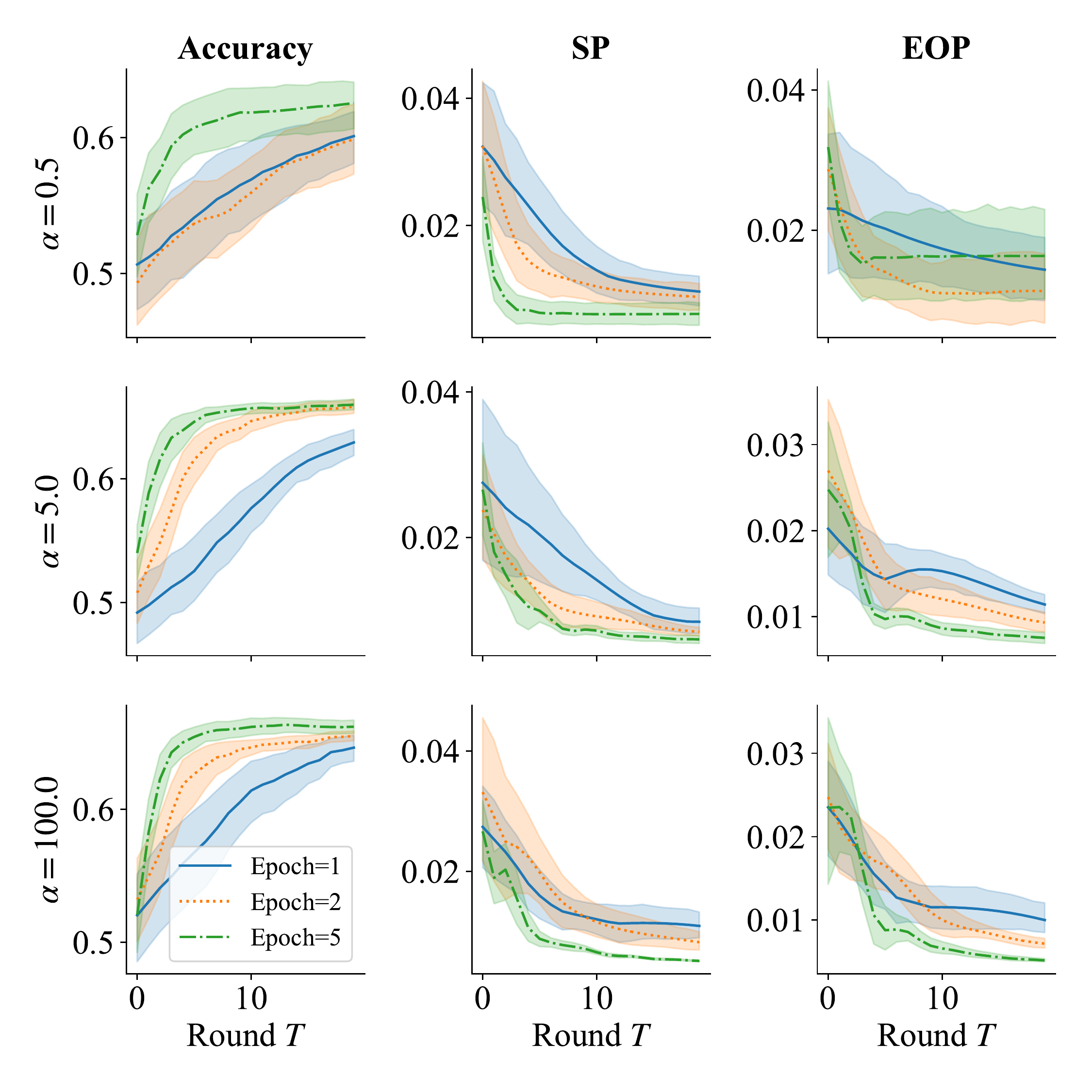}
        \caption{The accuracy and bias of the $\mfg$ method for different numbers of epochs $E$.}
        \label{fig:compas_epoch}
    \end{minipage}  
    \vspace{-0.1cm}
    \end{figure}

    Experiment results are summarized in \Autoref{tab:adult}.
    We can see that $\mfg$ has the smallest bias among almost all situations, while the accuracy drop by using $\mfg$ is negligible. Actually, the accuracy of $\mfg$ is within two standard errors compared to the best method, while the bias significantly decreases even in the highly heterogeneous case. 
    We also plot the trajectory of accuracy and bias during the training, as illustrated in \Autoref{fig:adult}. The shaded area indicates the $95\%$ confidence interval. 
    Without bias mitigation, the bias will increase for higher accuracy, as shown by `\mbox{FedAvg}'. The decreases in the biases by using `FairFed' and `\local' are significantly less than $\mfg$. The results on COMPAS and CelebA datasets are highly consistent, as detailed in the supplementary material.  

\subsection{Ablation study}

    We present the influence of $\mfg$'s hyper-parameters on the COMPAS dataset, though the results are similar on the other two datasets. The default values of hyper-parameters are chosen as $K=10$, $E=1$, $\eta=0.002$, and $\lambda=20$. 

    \textbf{Number of epochs.} We use epochs $E=\{1,2,5\}$, and the trajectory of the accuracy and bias are presented in \Autoref{fig:compas_epoch}, which indicates that $\mfg$ is not sensitive to $E$.
     
\begin{figure}[tb]
    \begin{minipage}[t]{.49\textwidth}
        \centering
        \includegraphics[width=\textwidth]{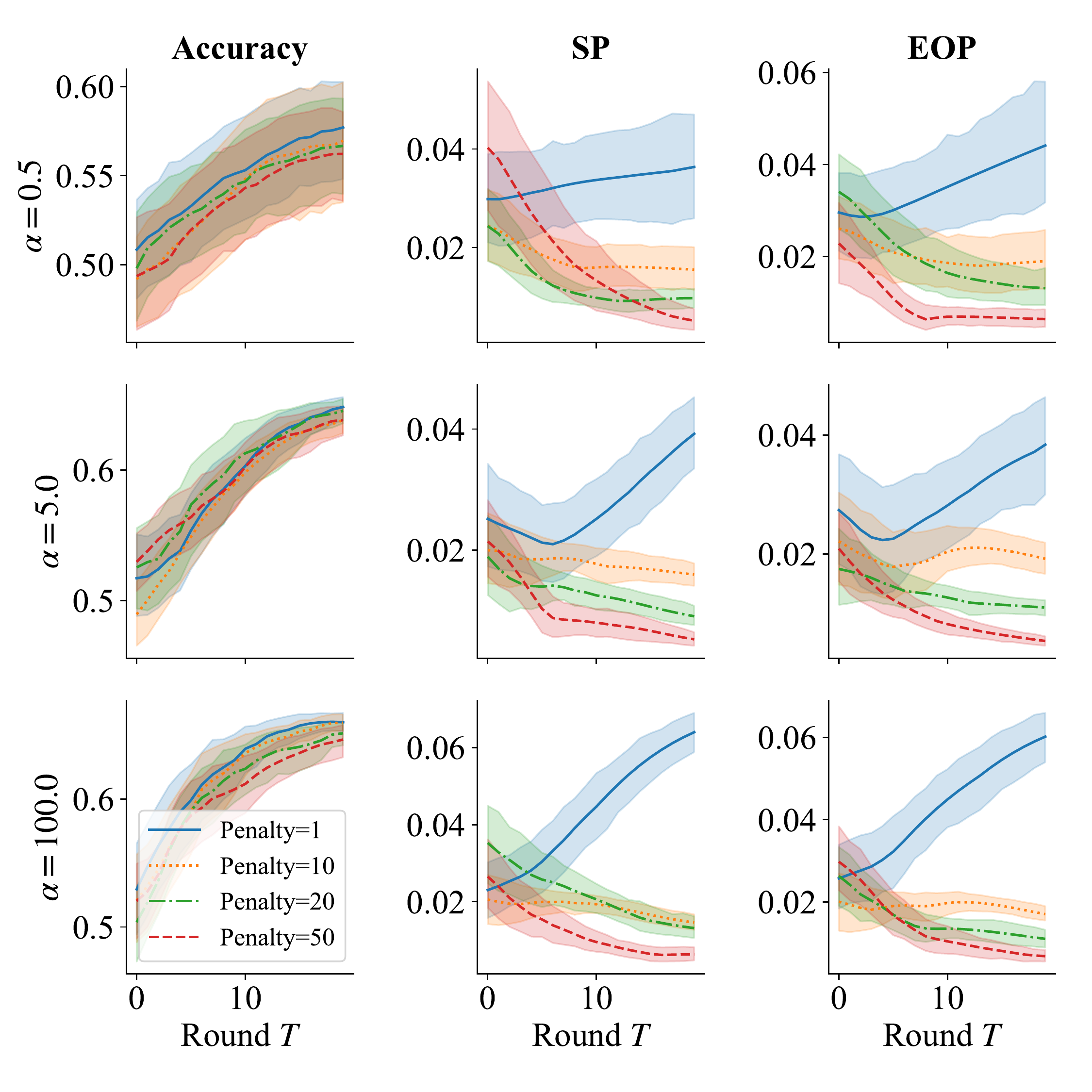}
        \caption{The accuracy and bias of $\mfg$ for different values of penalty parameter $\lambda$.}\label{fig:compas_pen}
    \end{minipage}
    \hfill
    \begin{minipage}[t]{.49\textwidth}
        \centering
        \includegraphics[width=\textwidth]{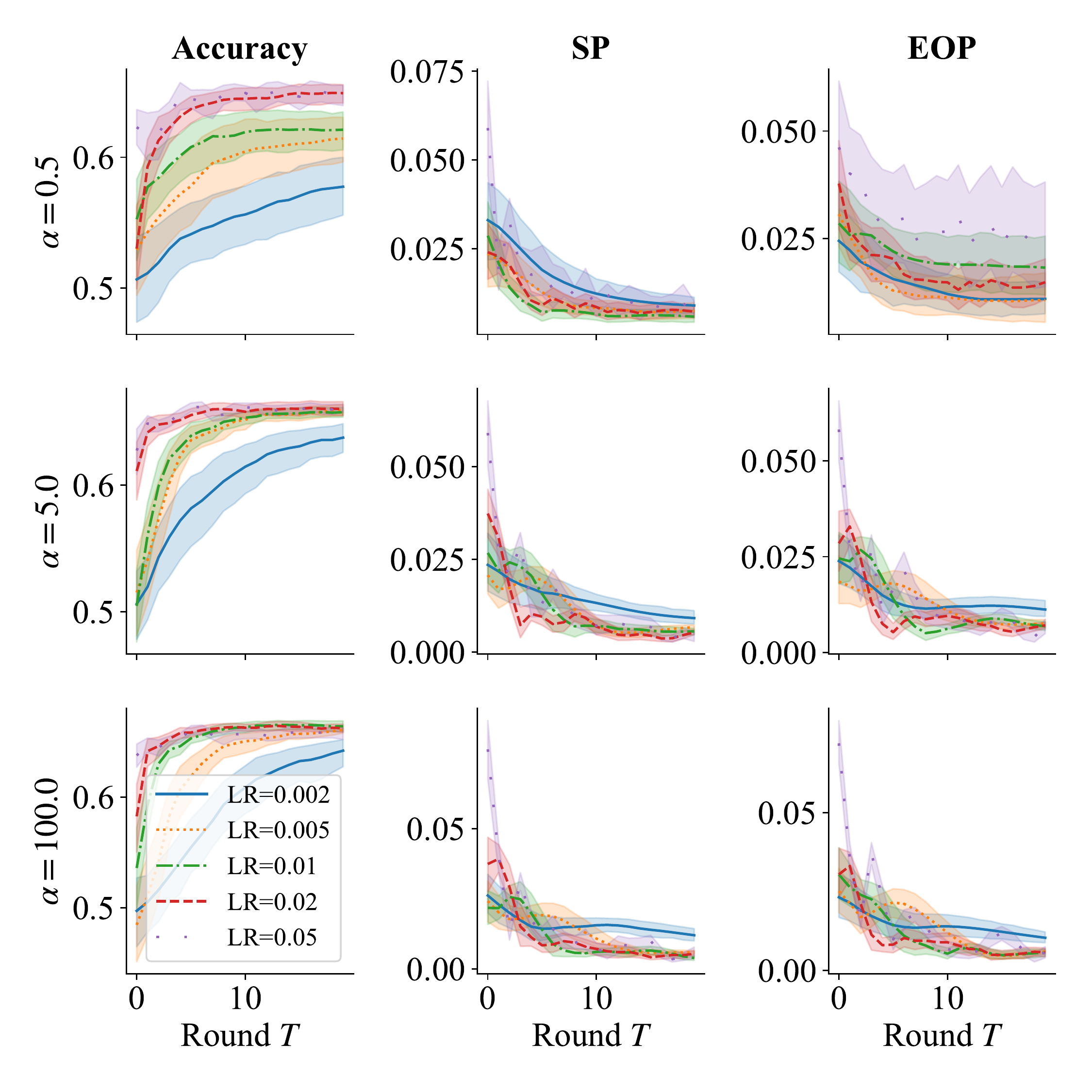}
        \caption{The accuracy and bias of $\mfg$ for different learning rate $\eta$.}
        \label{fig:compas_lr}
    \end{minipage}  
    \end{figure}

    \textbf{Penalty parameter.} We use the sequence $\lambda=\{1, 10, 20, 50\}$. \Autoref{fig:compas_pen} indicates that a higher penalty will enforce a smaller bias, while the accuracy decreases slightly.
    
    \textbf{Learning rate.} We choose $\eta=\{0.002$, $0.005$, $0.01$, $0.02$, $0.05\}$ and report the result in \Autoref{fig:compas_lr}. The result shows that a wide range of $\eta$ works well for $\mfg$. 


\section{Past works} \label{sec:past}
       
        For bias mitigation methods in federated learning, one popular approach is locally fair training as mentioned in the introduction. For example, \citet{abay2020mitigating, chu2021fedfair,ezzeldin2021fairfed} propose to train local models by applying centralized bias mitigation methods, such as reweighing the dataset to balance the group distribution~\citep{kamiran2012data}, and adding a constraint or a penalizing term of fairness on the optimization objective function~\citep{zafar2017fairness, kamishima2012fairness}.
        There have been a few works to understand locally fair training recently. \citet{zeng2021improving} showed that training locally fair models with federated learning is better than assembling locally fair models without iterative server-client updates, but worse than centralized training. \citet{chu2021fedfair} proved that for homogeneous clients 
        and a specific fairness metric, locally fair training yields a global model with a fairness guarantee. 
        
        Works on handling fairness in federated learning other than locally fair training have also been emerging. 
        \citet{mehrabi2022towards} assumed a validation dataset is available for evaluating the local fairness values and assigned higher weights to fairer clients. \citet{zhang2018mitigating} used a reinforcement learning approach to select clients that participate in the training with the highest local fairness and accuracy. \citet{du2021fairness} proposed to add a global fairness constraint to the agnostic federated learning formulation. \citet{rodriguez2021enforcing} proposed to solve a fairness-constrained optimization problem.  \citet{zeng2021improving} proposed to solve a bi-level optimization problem with the outer loop adaptively choosing fair batch representation of the training data. 
        In contrast with \citet{rodriguez2021enforcing, zeng2021improving}, our proposed algorithm \mfg is motivated from the developed theory on local and global fairness measures, considers the penalized optimization, and can be easily incorporated with most existing FL algorithms with one-line change of  client updating steps.
        
\section{Conclusion}
\label{s:conc}

In this work, we proved that the fairness of the global model in federated learning is upper-bounded by the fairness of local models and the data heterogeneity level for proper group-based fairness metrics, thus providing theoretical support for locally fair training methods. Nevertheless, locally fair training may fail in highly heterogeneous cases. We also proposed a globally fair training method called \mfg for proper group-based fairness metrics, which directly minimizes the fairness-penalized empirical loss of the global model and can be easily incorporated with existing FL algorithms. Experiments on three real-world datasets showed that the proposed method can significantly reduce the model bias while retaining a similar prediction accuracy compared to the baseline. 

\textbf{Limitations} There are several problems not fully addressed and will be interesting future work.
First, the calibration is not a proper fairness metric, thus, how to generalize our results to calibration, or more generally, group-based metrics, is of interest. Second, the proposed method can be generalized to a multi-class response and a multi-class group variable. 
It is also worth thinking about the fairness issue with respect to multiple group variables.

\bibliography{ref}

\begin{thebibliography}{10}
\providecommand{\url}[1]{#1}
\csname url@samestyle\endcsname
\providecommand{\newblock}{\relax}
\providecommand{\bibinfo}[2]{#2}
\providecommand{\BIBentrySTDinterwordspacing}{\spaceskip=0pt\relax}
\providecommand{\BIBentryALTinterwordstretchfactor}{4}
\providecommand{\BIBentryALTinterwordspacing}{\spaceskip=\fontdimen2\font plus
\BIBentryALTinterwordstretchfactor\fontdimen3\font minus
  \fontdimen4\font\relax}
\providecommand{\BIBforeignlanguage}[2]{{%
\expandafter\ifx\csname l@#1\endcsname\relax
\typeout{** WARNING: IEEEtran.bst: No hyphenation pattern has been}%
\typeout{** loaded for the language `#1'. Using the pattern for}%
\typeout{** the default language instead.}%
\else
\language=\csname l@#1\endcsname
\fi
#2}}
\providecommand{\BIBdecl}{\relax}
\BIBdecl

\bibitem{mcmahan2017communication}
B.~McMahan, E.~Moore, D.~Ramage, S.~Hampson, and B.~A. y~Arcas,
  ``Communication-efficient learning of deep networks from decentralized
  data,'' in \emph{Artificial intelligence and statistics}.\hskip 1em plus
  0.5em minus 0.4em\relax PMLR, 2017, pp. 1273--1282.

\bibitem{konevcny2016federated}
J.~Kone{\v{c}}n{\`y}, H.~B. McMahan, F.~X. Yu, P.~Richt{\'a}rik, A.~T. Suresh,
  and D.~Bacon, ``Federated learning: Strategies for improving communication
  efficiency,'' \emph{arXiv preprint arXiv:1610.05492}, 2016.

\bibitem{li2020federated}
T.~Li, A.~K. Sahu, M.~Zaheer, M.~Sanjabi, A.~Talwalkar, and V.~Smith,
  ``Federated optimization in heterogeneous networks,'' \emph{Proc. MLSys},
  vol.~2, pp. 429--450, 2020.

\bibitem{berk2019accuracy}
R.~Berk, ``Accuracy and fairness for juvenile justice risk assessments,''
  \emph{Journal of Empirical Legal Studies}, vol.~16, no.~1, pp. 175--194,
  2019.

\bibitem{barocas2016big}
S.~Barocas and A.~D. Selbst, ``Big data's disparate impact,'' \emph{California
  law review}, pp. 671--732, 2016.

\bibitem{becker2010economics}
G.~S. Becker, \emph{The economics of discrimination}.\hskip 1em plus 0.5em
  minus 0.4em\relax University of Chicago press, 2010.

\bibitem{guion1966employment}
R.~M. Guion, ``Employment tests and discriminatory hiring,'' \emph{Industrial
  Relations: A Journal of Economy and Society}, vol.~5, no.~2, pp. 20--37,
  1966.

\bibitem{cleary1968test}
T.~A. Cleary, ``Test bias: Prediction of grades of negro and white students in
  integrated colleges,'' \emph{Journal of Educational Measurement}, vol.~5,
  no.~2, pp. 115--124, 1968.

\bibitem{hutchinson201950}
B.~Hutchinson and M.~Mitchell, ``50 years of test (un) fairness: Lessons for
  machine learning,'' in \emph{Proc. FACCT}, 2019, pp. 49--58.

\bibitem{angwin2016machine}
J.~Angwin, J.~Larson, S.~Mattu, and L.~Kirchner, ``Machine bias,'' in
  \emph{Ethics of Data and Analytics}.\hskip 1em plus 0.5em minus 0.4em\relax
  Auerbach Publications, 2016, pp. 254--264.

\bibitem{kamiran2012data}
F.~Kamiran and T.~Calders, ``Data preprocessing techniques for classification
  without discrimination,'' \emph{Knowledge and information systems}, vol.~33,
  no.~1, pp. 1--33, 2012.

\bibitem{kamishima2012fairness}
T.~Kamishima, S.~Akaho, H.~Asoh, and J.~Sakuma, ``Fairness-aware classifier
  with prejudice remover regularizer,'' in \emph{Proc. ECML PKDD}.\hskip 1em
  plus 0.5em minus 0.4em\relax Springer, 2012, pp. 35--50.

\bibitem{zafar2017fairness}
M.~B. Zafar, I.~Valera, M.~G. Rogriguez, and K.~P. Gummadi, ``Fairness
  constraints: Mechanisms for fair classification,'' in \emph{Artificial
  Intelligence and Statistics}.\hskip 1em plus 0.5em minus 0.4em\relax PMLR,
  2017, pp. 962--970.

\bibitem{zhang2018mitigating}
B.~H. Zhang, B.~Lemoine, and M.~Mitchell, ``Mitigating unwanted biases with
  adversarial learning,'' in \emph{Proc. AIES}, 2018, pp. 335--340.

\bibitem{hardt2016equality}
M.~Hardt, E.~Price, and N.~Srebro, ``Equality of opportunity in supervised
  learning,'' \emph{Proc. NeurIPS}, vol.~29, 2016.

\bibitem{lohia2019bias}
P.~K. Lohia, K.~N. Ramamurthy, M.~Bhide, D.~Saha, K.~R. Varshney, and R.~Puri,
  ``Bias mitigation post-processing for individual and group fairness,'' in
  \emph{Proc. ICASSP}.\hskip 1em plus 0.5em minus 0.4em\relax IEEE, 2019, pp.
  2847--2851.

\bibitem{caton2020fairness}
S.~Caton and C.~Haas, ``Fairness in machine learning: A survey,'' \emph{arXiv
  preprint arXiv:2010.04053}, 2020.

\bibitem{abay2020mitigating}
A.~Abay, Y.~Zhou, N.~Baracaldo, S.~Rajamoni, E.~Chuba, and H.~Ludwig,
  ``Mitigating bias in federated learning,'' \emph{arXiv preprint
  arXiv:2012.02447}, 2020.

\bibitem{ezzeldin2021fairfed}
Y.~H. Ezzeldin, S.~Yan, C.~He, E.~Ferrara, and S.~Avestimehr, ``Fairfed:
  Enabling group fairness in federated learning,'' \emph{arXiv preprint
  arXiv:2110.00857}, 2021.

\bibitem{li2020federatedl}
T.~Li, A.~K. Sahu, A.~Talwalkar, and V.~Smith, ``Federated learning:
  Challenges, methods, and future directions,'' \emph{IEEE Signal Processing
  Magazine}, vol.~37, no.~3, pp. 50--60, 2020.

\bibitem{mehrabi2021survey}
N.~Mehrabi, F.~Morstatter, N.~Saxena, K.~Lerman, and A.~Galstyan, ``A survey on
  bias and fairness in machine learning,'' \emph{ACM Computing Surveys (CSUR)},
  vol.~54, no.~6, pp. 1--35, 2021.

\bibitem{corbett2018measure}
S.~Corbett-Davies and S.~Goel, ``The measure and mismeasure of fairness: A
  critical review of fair machine learning,'' \emph{arXiv preprint
  arXiv:1808.00023}, 2018.

\bibitem{feldman2015certifying}
M.~Feldman, S.~A. Friedler, J.~Moeller, C.~Scheidegger, and
  S.~Venkatasubramanian, ``Certifying and removing disparate impact,'' in
  \emph{Proc. SIGKDD}, 2015, pp. 259--268.

\bibitem{pleiss2017fairness}
G.~Pleiss, M.~Raghavan, F.~Wu, J.~Kleinberg, and K.~Q. Weinberger, ``On
  fairness and calibration,'' \emph{Proc. NeurIPS}, vol.~30, 2017.

\bibitem{kleinberg2016inherent}
J.~Kleinberg, S.~Mullainathan, and M.~Raghavan, ``Inherent trade-offs in the
  fair determination of risk scores,'' \emph{arXiv preprint arXiv:1609.05807},
  2016.

\bibitem{kim2020fact}
J.~S. Kim, J.~Chen, and A.~Talwalkar, ``Fact: A diagnostic for group fairness
  trade-offs,'' in \emph{International Conference on Machine Learning}.\hskip
  1em plus 0.5em minus 0.4em\relax PMLR, 2020, pp. 5264--5274.

\bibitem{chu2021fedfair}
L.~Chu, L.~Wang, Y.~Dong, J.~Pei, Z.~Zhou, and Y.~Zhang, ``Fedfair: Training
  fair models in cross-silo federated learning,'' \emph{arXiv preprint
  arXiv:2109.05662}, 2021.

\bibitem{blyth1972simpson}
C.~R. Blyth, ``On simpson's paradox and the sure-thing principle,''
  \emph{Journal of the American Statistical Association}, vol.~67, no. 338, pp.
  364--366, 1972.

\bibitem{bickel1975sex}
P.~J. Bickel, E.~A. Hammel, and J.~W. O'Connell, ``Sex bias in graduate
  admissions: Data from berkeley: Measuring bias is harder than is usually
  assumed, and the evidence is sometimes contrary to expectation.''
  \emph{Science}, vol. 187, no. 4175, pp. 398--404, 1975.

\bibitem{bellamy2019ai}
R.~K. Bellamy, K.~Dey, M.~Hind, S.~C. Hoffman, S.~Houde, K.~Kannan, P.~Lohia,
  J.~Martino, S.~Mehta, A.~Mojsilovi{\'c} \emph{et~al.}, ``Ai fairness 360: An
  extensible toolkit for detecting and mitigating algorithmic bias,'' \emph{IBM
  Journal of Research and Development}, vol.~63, no. 4/5, pp. 4--1, 2019.

\bibitem{berk2017convex}
R.~Berk, H.~Heidari, S.~Jabbari, M.~Joseph, M.~Kearns, J.~Morgenstern, S.~Neel,
  and A.~Roth, ``A convex framework for fair regression,'' \emph{arXiv preprint
  arXiv:1706.02409}, 2017.

\bibitem{Dua:2019}
\BIBentryALTinterwordspacing
D.~Dua and C.~Graff, ``{UCI} machine learning repository,'' 2017. [Online].
  Available: \url{http://archive.ics.uci.edu/ml}
\BIBentrySTDinterwordspacing

\bibitem{liu2015faceattributes}
Z.~Liu, P.~Luo, X.~Wang, and X.~Tang, ``Deep learning face attributes in the
  wild,'' in \emph{Proc. ICCV}, December 2015.

\bibitem{he2016deep}
K.~He, X.~Zhang, S.~Ren, and J.~Sun, ``Deep residual learning for image
  recognition,'' in \emph{Proceedings of the IEEE conference on computer vision
  and pattern recognition}, 2016, pp. 770--778.

\bibitem{zeng2021improving}
Y.~Zeng, H.~Chen, and K.~Lee, ``Improving fairness via federated learning,''
  \emph{arXiv preprint arXiv:2110.15545}, 2021.

\bibitem{mehrabi2022towards}
N.~Mehrabi, C.~de~Lichy, J.~McKay, C.~He, and W.~Campbell, ``Towards
  multi-objective statistically fair federated learning,'' \emph{arXiv preprint
  arXiv:2201.09917}, 2022.

\bibitem{du2021fairness}
W.~Du, D.~Xu, X.~Wu, and H.~Tong, ``Fairness-aware agnostic federated
  learning,'' in \emph{Proc. SDM}.\hskip 1em plus 0.5em minus 0.4em\relax SIAM,
  2021, pp. 181--189.

\bibitem{rodriguez2021enforcing}
B.~Rodr{\'\i}guez-G{\'a}lvez, F.~Granqvist, R.~van Dalen, and M.~Seigel,
  ``Enforcing fairness in private federated learning via the modified method of
  differential multipliers,'' \emph{arXiv preprint arXiv:2109.08604}, 2021.

\bibitem{li2020fedopt}
T.~Li, A.~K. Sahu, M.~Zaheer, M.~Sanjabi, A.~Talwalkar, and V.~Smith,
  ``Federated optimization in heterogeneous networks,'' \emph{Proceedings of
  Machine Learning and Systems}, vol.~2, pp. 429--450, 2020.

\bibitem{wang2020tackling}
J.~Wang, Q.~Liu, H.~Liang, G.~Joshi, and H.~V. Poor, ``Tackling the objective
  inconsistency problem in heterogeneous federated optimization,'' \emph{Proc.
  NeurIPS}, vol.~33, pp. 7611--7623, 2020.

\end{thebibliography}
\bibliographystyle{IEEEtran}

\newpage
\appendix
\section{Fairness metrics}
In this appendix section, we validate that three fairness metrics (SP, EOP, and Calibration) satisfy our Definition~\ref{def:fair}. Furthermore, SP and EOP are proper group-based fairness metric Definition~\ref{def:proper}. 

\textbf{Statistical Parity.} Recall that $F(f, \mathcal{D}) = \abs{\P(\hat{Y}=1|A=0)-\P(\hat{Y}=1|A=1)}$. By Bayes' Theorem, we have 
\begin{align*}
    \P(\hat{Y}=1|A=0) = \frac{\P(\hat{Y}=1, A=0)}{\P(A=0)}. 
\end{align*}
Therefore, let $a(f, \mathcal{D}) = \P(\hat{Y}=1, A=0)$, $b(f, \mathcal{D}) = \P(A=0)$, then $\P(\hat{Y}=1|A=0) = a(f, \mathcal{D})/b(f, \mathcal{D})$. Similarly, we have $\P(\hat{Y}=1|A=1) = c(f, \mathcal{D})/d(f, \mathcal{D})$, where $c(f, \mathcal{D}) = \P(\hat{Y}=1, A=1)$, $d(f, \mathcal{D}) = \P(A=1)$. Thus, SP is a group-based fairness metric. Furthermore, it is clear that $b(f, \mathcal{D})$ and $d(f, \mathcal{D})$ are independent of $f$, hence SP is also a proper group-based fairness metric. 

\textbf{Equal Opportunity.} For EOP, $F(f, \mathcal{D}) = \abs{\P(\hat{Y}=1|A=0,Y=1)-\P(\hat{Y}=1|A=1,Y=1)}$. Using Bayes' Theorem again, we know
\begin{align*}
    \P(\hat{Y}=1|A=0,Y=1) = \frac{\P(\hat{Y}=1,A=0,Y=1)}{\P(A=0,Y=1)}, \ \P(\hat{Y}=1|A=1,Y=1) = \frac{\P(\hat{Y}=1,A=1,Y=1)}{\P(A=1,Y=1)}, 
\end{align*}
which aligns with Table~\ref{tab:abcd}.
    
\textbf{Well-Calibration.} In this case, $F(f, \mathcal{D}) = \abs{\P(Y=1|A=0,\hat{Y}=1)-\P(Y=1|A=1,\hat{Y}=1)}$, and
\begin{align*}
    \P(Y=1|A=0,\hat{Y}=1) = \frac{\P(Y=1,A=0,\hat{Y}=1)}{\P(A=0,\hat{Y}=1)}, \P(Y=1|A=1,\hat{Y}=1) = \frac{\P(Y=1,A=1,\hat{Y}=1)}{\P(A=1,\hat{Y}=1)}.
\end{align*}
We note that for calibration, $b(f, \mathcal{D}) = \P(A=0,\hat{Y}=1)$ and $d(f, \mathcal{D}) = \P(A=1,\hat{Y}=1)$, which are functions of both function $f$ and distribution $\mathcal{D}$.

\section{Missing Proofs in Section 4}
\textbf{Proof of Theorem~\ref{thm:general}.} 
We first prove that fair local models do not imply a fair global model. Let $g_k$ and $g$ be the abbreviations of $g(f, \hat{\mathcal{D}}_k)$ and $g(f, \hat{\mathcal{D}})$ for $g \in \{a, b, c, d\}$ (see Definition~\ref{def:fair}), respectively. Since all $a,b,c,d$ are expectations, they are linear in the data distribution by the property of expectation. 
Thus, we only need to show that there exist a $f$ and data distributions $\hat{\mathcal{D}}_k$'s such that 
\begin{align*}
    F(f, \hat{\mathcal{D}}) = F(f, \sum_k w_k \hat{\mathcal{D}}_k) = \biggl|  \frac{\sum_k w_k a_k}{\sum_k w_k b_k} - \frac{\sum_k w_k c_k}{\sum_k w_k d_k} \biggr| \geq C, \numberthis \label{eq:thm1step1}
\end{align*}
where $w_k=n_k/n$. 
Note that $w_k$ can take an arbitrary value in $[0,1]$ as long as $\hat{\mathcal{D}}_k$'s are properly chosen. Furthermore, according to Definition~\ref{def:fair}, $g_k$'s are arbitrarily manipulable as well. 
Next, we will construct $g_k$'s and $w_k$'s that satisfy Eq.~\ref{eq:thm1step1}. 
In particular, we consider quantities with $a_1/b_1=c_1/d_1 = 1$, $w_1=(1+C)/2$, and $a_k/b_k=c_k/d_k=0$ and $w_k=(1-w_1)/(K-1)$ for $k=2,\dots,K$. By simple calculation, when $b_1, d_2, \dots, d_K$ converge to one and $d_1, b_2, \dots, b_K$ converge to zero, $F(f, \hat{\mathcal{D}})$ converges to $w_1$, which is larger than $C$. It immediately implies that there exists a proper choice satisfying Eq.~\ref{eq:thm1step1}.

As for the converse result, the following choice suffices: 
\begin{align*}
     & a_{2l}/b_{2l} = C, a_{2l+1}/b_{2l+1} = 0,
    c_{2l}/d_{2l} = 0, c_{2l+1}/d_{2l+1} = C,
    \\& w_{2l} = \frac{\lfloor (K+1)/2 \rfloor}{2\lfloor K/2 \rfloor\lfloor (K+1)/2 \rfloor}, w_{2l+1} = \frac{\lfloor K/2 \rfloor}{2\lfloor K/2 \rfloor\lfloor (K+1)/2 \rfloor},  l=0,\dots, \lfloor K/2 \rfloor,
\end{align*}
where $\lfloor x \rfloor$ means the floor of a number $x$. 

\textbf{Proof of Corollary~\ref{coro:general}.} If the claim is false, then there exists a sequence of constants $\{v_k, k=1,\dots,K\}$, such that $F(f, \hat{\mathcal{D}}) = \sum_{k=1}^K v_k F(f, \hat{\mathcal{D}}_k)$ always holds. Now, evoking Theorem~\ref{thm:general}, we know it is possible that $F(f, \hat{\mathcal{D}}) >0 $ with $F(f, \hat{\mathcal{D}}_k)=0$ for all $k$, which is a contradiction, and thus concludes the proof.

\textbf{Proof of Theorem~\ref{thm:perfect_fairness}.} Recall that $F(f,\mathcal{D}_k) = \abs{a_k/b_k-c_k/d_k}$. From Eq.~\ref{eq:thm1step1}, we know that 
\begin{align*}
    F(f, \mathcal{D})=0 \Longleftrightarrow \frac{\sum_k w_ka_k}{\sum_k w_kb_k} = \frac{\sum_k w_kc_k}{\sum_k w_kd_k}.
\end{align*}
Multiplying $(\sum_k w_kb_k)(\sum_k w_kd_k)$ on the both hand sides and rearranging the above equation yields that $\w^\T M \w = 0$, where $\w = (w_1,\dots,w_K)^\T$ and $M\in \real^{K \times K}$ is a matrix with $(i,j)$-th element $M_{ij} = a_id_j-b_ic_j$. 
Thus, $\w^\T M \w = 0$ for any $\w$ is equivalent to that $a_id_j-b_ic_j = 0$ for all $1\leq i,j \leq K$, which completes the proof. 

\textbf{Proof of Theorem~\ref{thm:neariid}.}
The local fairness condition $F(f, \hat{\mathcal{D}}_k) \leq \alpha$ gives that $\abs{a_k/b_k - c_k/d_k} \leq \alpha,$ thus $a_k \leq (\alpha + c_k/d_k)b_k$, and we have
        \begin{align*}
             \frac{\sum_k w_ka_k}{\sum_k w_kb_k} - \frac{\sum_k w_kc_k}{\sum_k w_kd_k} 
            & \leq \alpha + \frac{\sum_k w_k c_k b_k/d_k}{\sum_k w_kb_k} - \frac{\sum_k w_kc_k}{\sum_k w_kd_k} 
            \\& \leq \alpha + \frac{\sum_k w_k c_k}{\sum_k w_kd_k} \biggl(\frac{d}{b}\frac{b_k}{d_k}-1\biggr) \leq \alpha+\beta.
        \end{align*}
        The last step is due to $c_k/d_k \leq 1$ and the definition of data heterogeneity coefficient. Similarly, we have $a_k \geq (c_k/d_k-\alpha)b_k$ and 
        \begin{align*}
            \frac{\sum_k w_ka_k}{\sum_k w_kb_k} - \frac{\sum_k w_kc_k}{\sum_k w_kd_k} &\geq \frac{\sum_k w_k c_k b_k/d_k}{\sum_k w_kb_k} - \alpha - \frac{\sum_k w_kc_k}{\sum_k w_kd_k} 
            \\& \geq -\alpha + \frac{\sum_k w_k c_k}{\sum_k w_kd_k} \biggl(\frac{d}{b}\frac{b_k}{d_k}-1\biggr) \geq -(\alpha+\beta),
        \end{align*}
        which concludes the proof.

\section{Missing Proofs in Section 5}\label{appendix_convergence}
\newcommand{\de}{\define}
\subsection{Proof of Theorem~\ref{thm:gradient}.}
     For a proper group-based fairness metric $F$, we have
        \begin{align*}
            F(f, \hat{\mathcal{D}}_k) &= \biggl | \frac{a_k}{b_k}-\frac{c_k}{d_k} \biggr |, 
            \\ F(f, \hat{\mathcal{D}}) &=  \biggl | \frac{\sum_k w_ka_k}{\sum_k w_kb_k} - \frac{\sum_k w_kc_k}{\sum_k w_kd_k} \biggr |,
        \end{align*}
        where $a_k$ and $c_k$ are functions of $f$ and $\hat{\mathcal{D}}_k$, and $b_k$ and $d_k$ are functions of $\hat{\mathcal{D}}_k$ only. Recall that $b = \sum_k w_kb_k$, $d =\sum_k  w_kd_k$, and $F_k = a_k/b-c_k/d$, it is straightforward to verify that 
        \begin{align*}
            F(f, \hat{\mathcal{D}}) = \abs{\tilde{F}}, \ \tilde{F} = \frac{\sum_k w_ka_k}{\sum_k w_kb_k} - \frac{\sum_k w_kc_k}{\sum_k w_kd_k} = \sum_k w_k F_k.
        \end{align*}
        Therefore,
        \begin{align*}
            \nabla_{\theta} F(f,\hat{\mathcal{D}})
            &= \sign(\tilde{F}(f,\hat{\mathcal{D}})) \biggl( \frac{\sum_k w_k \nabla_{\theta}a_k}{\sum_k w_kb_k} - \frac{\sum_k w_k \nabla_{\theta}c_k}{\sum_k w_kd_k} \biggr)
            \\ &=  \sign(\tilde{F}(f,\hat{\mathcal{D}})) \sum_{k=1}^K w_k \biggl( \frac{ \nabla_{\theta}a_k}{b} - \frac{ \nabla_{\theta}c_k}{d} \biggr)
            \\ &= \sum_{k=1}^K w_k \sign(\tilde{F}(f,\hat{\mathcal{D}})) \nabla_{\theta} F_k.
        \end{align*}
        Finally, by chain rule, we have that
        \begin{align*}
        \nabla_{\theta} J(F(f,\hat{\mathcal{D}}))&=   \nabla_{F} J(F(f,\hat{\mathcal{D}})) \nabla_{\theta} F(f,\hat{\mathcal{D}})
        \\& =\sign(\tilde{F}) \nabla_F J(F(f,\hat{\mathcal{D}}))  \sum_{k=1}^K w_k \nabla_{\theta} F_k, 
        \end{align*}
        which completes the proof.    

    \subsection{Convergence analysis}
    We first restate the problem setup for clarity. 
    Recall the global objective function Eq.~\eqref{eq:obj_pen} is
    \begin{align*}
        \min_{\theta} L(\theta) = \sum_{k=1}^K \frac{n_k}{n} L_k(\theta) + \lambda J(F(f(\cdot ;\theta); \hat{\mathcal{D}})).
    \end{align*}
    And the local objective functions are
    \begin{align*}
        \min_{\theta} H_k(\theta) \de L_k(\theta)+\lambda C_{\theta^{t-1}} F_k(\theta).
    \end{align*}
    Next, we state the training procedure with random client selection and stochastic gradient descent optimization. In particular, at the communication round $t$, we have
    \begin{align*}
        \theta^{t+1}_k &= \theta^{t} - \eta_t g_k(\theta^{t} \mid \xi), k \in S_t, \\ 
        \theta^{t+1} &= \frac{1}{K} \sum_{k \in S_t} \theta^{t+1}_k,
    \end{align*}
    where $g_k(\theta^t\mid \xi)$ is the stochastic gradient of $H_k$, $\xi$ represents the stochastic batches of datasets, $S_t$ is a randomly selected subset of clients with cardinality $M$ (in which client $k$ is selected with probability $n_k/n$), and $\eta_t$ is the step~size.

    We make the following technical assumptions often used in the optimization literature, e.g.,~\cite{li2020fedopt, wang2020tackling} and the references therein. For two vectors $u$ and $v$, $\ip{u}{v}=u^\T v$ is the inner product of $u$ and $v$, and $\norm{v}=(v^\T v)^{1/2}$ is the $\ell_2$ norm of $v$. The gradient operator $\nabla$ is with respect to the model parameter $\theta$ throughout this subsection. 
    
    \begin{assumption}[Smoothness]\label{asmp:smooth}
        The gradients of $L_k$'s and $J$ are $L$-Lipshitz continuous. A function $f(\cdot)$ is $L$-Lipshitz continuous if for any $x, y$ we have
        \begin{align*}
            \norm{\nabla f(x) - \nabla f(y)} \leq L \norm{x-y}.
        \end{align*}
    \end{assumption}
    \begin{assumption}[Unbiasedness]\label{asmp:unbias} The stochastic gradient is unbiased for all clients, that is, 
        $\E_\xi \{g_k(\theta^t\mid \xi)\} = \nabla H_k(\theta^t)$, for all $k=1,\dots, K$. 
    \end{assumption}
    \begin{assumption}[Bounded variance]\label{asmp:bounded} The stochastic gradient has a bounded variance for all clients, 
        namely $\E_\xi [\{g_k(\theta^t \mid \xi)-\nabla H_k(\theta^t)\}^2  ] \leq \sigma^2$, for all $ k=1,\dots, K$. 
    \end{assumption}
    \begin{assumption}[Bounded dissimilarity]\label{asmp:similar}
        There exist a constant $B\geq 1$ such that for all $\sum_{k=1}^K w_k =1, w_k \geq 0$, we have
        \begin{align*}
            \sum_{k=1}^K w_k \norm{\nabla H_k(\theta)}^2 \leq B^2 \biggl\|\sum_{k=1}^K w_k \nabla H_k(\theta)\biggr\|^2.
        \end{align*}
    \end{assumption}
    \begin{assumption}\label{asmp:low}
        The objective function is lower bounded, $L^* \de \inf_{\theta} L(\theta) > -\infty$.
    \end{assumption}
    
    \begin{remark}
        Assumptions~\ref{asmp:smooth},~\ref{asmp:unbias}, and~\ref{asmp:bounded} are standard in optimization literature, which ensure that the SGD update produces a sufficiently large decrease in the function value, leading to the convergence. 
        Assumption~\ref{asmp:similar} ensures the convergence with data heterogeneity. Larger $B$ indicate more severe data heterogeneity, and $B=1$ corresponds to the homogeneous case.
    \end{remark}
    \begin{remark}
        If we use GD instead of SGD, then the update of local clients will be
        \begin{align*}
            \theta^{t+1}_k &= \theta^{t+1}_l - \eta_t \nabla H_k(\theta^{t}), k \in S_t,
        \end{align*}
        and the Assumptions~\ref{asmp:unbias} and \ref{asmp:bounded} are automatically satisfied with $\sigma = 0$.
    \end{remark}
    
    Now, we restate and prove Theorem~\ref{thm:converge} below.

    \textbf{Theorem~\ref{thm:converge} (Convergence result)}
    Under Assumptions~\ref{asmp:smooth}-\ref{asmp:low}, when the step-size sequence $\{\eta_t, t=0,\dots,T-1 \}$ satisfies $C_0 \geq \eta_0 \geq \eta_t > 0$, we have 
    \begin{align*}
        \min_{t=0,\dots,T-1} \E ( \norm{\nabla L(\theta^t)}^2 ) \leq C\biggl(\frac{1}{\sum_{t=0}^{T-1} \eta_t} + \frac{ \sum_{t=0}^{T-1}\eta_t^2}{\sum_{t=0}^{T-1} \eta_t}\biggr),
    \end{align*}
    where $C_0$ and $C$ are two constants independent of $T$ and $\{\eta_t, t=0,\dots,T-1\}$.

    \renewcommand{\hat}{\widehat}  
    \renewcommand{\tilde}{\widetilde}
    \renewcommand{\bar}{\overline}

    \textbf{Proof of Theorem~\ref{thm:converge}}
        We assume $\theta^t$ is fixed for now and denote $ \hat{\theta}^{t+1} \de \theta^t-\eta_t \nabla L(\theta^t)$.
        Note that $\E (\theta^{t+1}) = \hat{\theta}^{t+1}$ by Theorem~\ref{thm:gradient}. 
        By Assumption~\ref{asmp:smooth}, we have 
        \begin{align*}
            \E\{L(\theta^{t+1})\} &\leq L(\theta^t) + \E\{\ip{\nabla L(\theta^t)}{\theta^{t+1} - \theta^t}\} + \E \biggl( \frac{L}{2}\norm{\theta^{t+1} - \theta^t}^2 \biggr)
            \\ &\leq L(\theta^t) +  \ip{\nabla L(\theta^t)}{\hat{\theta}^{t+1} - \theta^t} + L \bigl\{\norm{\hat{\theta}^{t+1} - \theta^t}^2 
            + \E ( \norm{\hat{\theta}^{t+1} - \theta^{t+1}}^2) \bigr\}
            \\ &=  L(\theta^t) -\eta_t(1-L\eta_t)\norm{\nabla L(\theta^t)}^2 + L \E ( \norm{\hat{\theta}^{t+1} - \theta^{t+1}}^2).
            \numberthis \label{eq:main}
        \end{align*}
        Since all clients are independent of each other, we have
        \begin{align*}
            \E(\norm{\hat{\theta}^{t+1} - \theta^{t+1}}^2) &=  \E_{\xi} \{\E_{S_t}(\norm{\hat{\theta}^{t+1} - \theta^{t+1}}^2)\}
            \\ &\leq \E_{\xi}\biggl\{ \frac{1}{M}\E_k(\norm{\theta^{t+1}_k-\hat{\theta}^{t+1}}^2) \biggr\}
            \\ &= \frac{\eta_t^2}{M} \E_{\xi}\biggl\{\E_k(\norm{g_k(\theta^{t} \mid \xi)-\nabla L(\theta^{t})}^2)\biggr\}
            \\\text{(triangle inequality and Assumption~\ref{asmp:bounded})} &\leq \frac{2\eta_t^2}{M} \biggl\{\E_k(\norm{\nabla H_k(\theta^{t})-\nabla L(\theta^{t})}^2) +\sigma^2 \biggr\}
            \\\text{(triangle inequality)} &\leq \frac{4\eta_t^2}{M} \biggl\{\E_k(\norm{\nabla H_k(\theta^{t})}^2) + \norm{\nabla L(\theta^{t})}^2 +\sigma^2 \biggr\}
            \\ \text{(Assumption~\ref{asmp:similar})}&\leq \frac{4\eta_t^2}{M} \biggl\{(B^2+1)\norm{\nabla L(\theta^{t})}^2  +\sigma^2\biggr\}. \numberthis\label{theta_diff_1}
        \end{align*}
        Plugging Eqs.~\eqref{theta_diff_1} into Eq.~\eqref{eq:main}, we have 
        \begin{align*}
            \E\{L(\theta^{t+1})\} &\leq L(\theta^t) -\eta_t(1-L\eta_t)\norm{\nabla L(\theta^t)}^2 +4M^{-1}L\eta_t^2 \{(B^2+1) \norm{\nabla L(\theta^t)}^2  + \sigma^2\}
            \\ &= L(\theta^t) -\eta_t[1-L\eta_t\{1+4M^{-1}L(B^2+1) \}]\norm{\nabla L(\theta^t)}^2 +4M^{-1}L\eta_t^2 \sigma^2
             \\ &= L(\theta^t) -\eta_t(1-c_1 \eta_t)\norm{\nabla L(\theta^t)}^2 + c_2 \eta_t^2, \numberthis\label{eq:intermedia}
        \end{align*}
        where $c_1=L\{1+4M^{-1}L(B^2+1) \}$ and $c_2=4M^{-1}L\sigma^2$.
        Next, we take expectation on Eq.~\eqref{eq:intermedia}, reorganize and sum it from $t=0$ to $t=T-1$, obtaining
        \begin{align*}
            \sum_{t=0}^{T-1} \eta_t(1-c_1 \eta_t) \E \bigl(\norm{\nabla L(\theta^t)}^2 \bigr) \leq \E\{L(\theta^0)-L(\theta^{t+1})\}+ \sum_{t=0}^{T-1} c_2 \eta_t^2.
        \end{align*}
        For a sufficiently small $\eta_t$-sequence such that $\eta_t \leq 1/(2c_1)$ for all $t$,
        we have
        \begin{align*}
            \min_{t=0, \dots, T-1}  \E \bigl(\norm{\nabla L(\theta^t)}^2 \bigr) \sum_{t=0}^{T-1} \eta_t /2 \leq  \sum_{t=0}^{T-1}\frac{\eta_t}{2}  \E \bigl(\norm{\nabla L(\theta^t)}^2 \bigr) \leq \E\{L(\theta^0)-L(\theta^{T})\}+ \sum_{t=0}^{T-1} c_2 \eta_t^2.
        \end{align*}
        As a result,
        \begin{align*}
            \min_{t=0, \dots, T-1} \E \bigl(\norm{\nabla L(\theta^t)}^2 \bigr) \leq \frac{2\E\{L(\theta^0)-L^*\}}{\sum_{t=0}^{T-1} \eta_t} + \frac{2c_2 \sum_{t=0}^{T-1}\eta_t^2}{\sum_{t=0}^{T-1} \eta_t},
        \end{align*}    
        which concludes the proof.

    \textbf{Proof of Corollary~\ref{coro:covergence}.}
    The first two statements regarding the special choices of the step-size sequence $\eta_t$ are directly obtained from Theorem~\ref{thm:converge}. As for the gradient descent, when $\sigma=0$, the constant $c_2$ in the proof of Theorem~\ref{thm:converge} disappears. This leads to 
    \begin{align*}
        \min_{t=0, \dots, T-1} \E \bigl(\norm{\nabla L(\theta^t)}^2 \bigr) \leq \frac{2\E\{L(W_0)-L^*\}}{\sum_{t=0}^{T-1} \eta_t},
    \end{align*}
    which completes the proof.
    
\section{Further Experiments}

\SetKwInput{Input}{Input}
\SetKwProg{kwSystem}{System executes}{:}{}
\SetKwProg{kwServer}{SignUpdate}{:}{}
\SetKwProg{kwClient}{ClientUpdate}{:}{}

\subsection{Algorithms used in experiments}
For completeness, we state the algorithms of `FedAvg', `FairFed', and `LFT' in Algorithm~\ref{alg:fedavg}, Algorithm~\ref{alg:fairfed}, and Algorithm~\ref{alg:lft}, respectively.

\begin{algorithm}[H]
        \SetAlgoLined
\DontPrintSemicolon
        \caption{(`FedAvg') Federated Average}\label{alg:fedavg}
            \Input{Communication rounds $T$, learning rate $\eta$, local training epochs $E$.}
            \kwSystem{}{
            Initialize the global model parameters $\theta^0$
            
    \For{\textup{each communication round }$t = 1,2,\dots T$}{
        \For{\textup{each client }$k =1,\dots, K$\textup{ \textbf{in parallel}}}{
    
            Receive the model parameters $\theta_k^{t,0} = \theta^{t-1}$ from the server
    
            $\theta_k^{t,E} \leftarrow$ \textbf{ClientUpdate}$(\theta_k^{t,0}, Z)$
        }
        
        Server update global model $\theta^t \leftarrow \sum_k w_k \theta_k^{t,E}$.
    }
    Return the final global model $f(\cdot;\theta^T)$
}
\kwClient{\textup{$(\theta_k^{t,0}, Z)$}}{
    \For{\textup{each local epoch $e$ from 1 to $E$}}{
    Perform gradient descent $\theta_k^{t,e} \leftarrow \theta_k^{t,e-1}-\eta \nabla_{\theta_k^{t,e-1}} L_k$
        }
    Return $\theta_k^{t,E}$ 
}
\end{algorithm}    

\begin{algorithm}[H]
        \SetAlgoLined
\DontPrintSemicolon
        \caption{(`FairFed') Fairness-aware Federated Average}\label{alg:fairfed}
            \Input{Communication rounds $T$, learning rate $\eta$, local training epochs $E$, hyper-parameter $\beta$.}
            \kwSystem{}{
            Initialize the global model parameters $\theta^0$
            
    \For{\textup{each communication round }$t = 1,2,\dots T$}{
        \For{\textup{each client }$k =1,\dots, K$\textup{ \textbf{in parallel}}}{
    
            Receive the model parameters $\theta_k^{t,0} = \theta^{t-1}$ from the server
    
            $\theta_k^{t,E}, F_k^t, m_k^t \leftarrow$ \textbf{ClientUpdate}$(\theta_k^{t,0}, Z)$
        }

        Calculate global fairness $F^t \leftarrow \sum_k w_k m_k^t$

        Aggregation weights $w_k^t \leftarrow \exp(-\beta \abs{F^t-F^t_k})w_k$
        
        Server update global model $\theta^t \leftarrow \sum_k w_k^t \theta_k^{t,E}$.
    }
    Return the final global model $f(\cdot;\theta^T)$
}
\kwClient{\textup{$(\theta_k^{t,0}, Z)$}}{
    \For{\textup{each local epoch $e$ from 1 to $E$}}{
    Perform any bias mitigation algorithm to this local client
        }
    Calculate the local fairness $F_k^t \leftarrow F(f(\cdot;\theta_k^{t,E}), \hat{\mathcal{D}}_k)$

    Calculate the global fairness component $m_k^t$ \tcc{See~\citet{ezzeldin2021fairfed}}
    
    Return $\theta_k^{t,E}, F_k^t, m_k^t$ 
}
\end{algorithm}    

    
    
        
\begin{algorithm}[H]
        \SetAlgoLined
\DontPrintSemicolon
        \caption{(`LRW') Locally reweighing}\label{alg:lft}
            \Input{Communication rounds $T$, learning rate $\eta$, local training epochs $E$, penalty parameter $\lambda$.}
            \kwSystem{}{
            Initialize the global model parameters $\theta^0$
    \For{\textup{each communication round }$t = 1,2,\dots T$}{
        \For{\textup{each client }$k =1,\dots, K$\textup{ \textbf{in parallel}}}{
    
            Receive the model parameters $\theta_k^{t,0} = \theta^{t-1}$ from the server
    
            $\theta_k^{t,E} \leftarrow$ \textbf{ClientUpdate}$(\theta_k^{t,0}, Z)$
        }
        
        Server update global model $\theta^t \leftarrow \sum_k w_k \theta_k^{t,E}$.
    }
    Return the final global model $f(\cdot;\theta^T)$
}
\kwClient{\textup{$(\theta_k^{t,0}, Z)$}}{
    \For{\textup{each local epoch $e$ from 1 to $E$}}{
    Assign each data point a score associated with its sensitive attribute \tcc{See~\citet{abay2020mitigating}}
    
    Perform ordinary gradient descent on the weighted loss function 
        }
    Return $\theta_k^{t,E}$ 
}
\end{algorithm}    

\subsection{Details of training}
The hyper-parameters used in Section 6.2 are summarized in \Autoref{tab:hyper}. The regularization function is $J(x)=x$. 
\begin{table}[H]
\centering
\caption{Hyper-parameters used in our experiments.}
\label{tab:hyper}
\begin{tabular}{@{}cccccccc@{}}
\toprule
\multicolumn{2}{c}{Dataset} & \multicolumn{2}{c}{Adult} & \multicolumn{2}{c}{COMPAS} & \multicolumn{2}{c}{CelebA} \\ \midrule
\multicolumn{2}{c}{Architecture} & \multicolumn{2}{c}{Linear} & \multicolumn{2}{c}{Linear} & \multicolumn{2}{c}{ResNet18} \\ \midrule

\multicolumn{2}{c}{Number of clients} & \multicolumn{2}{c}{10} & \multicolumn{2}{c}{10}& \multicolumn{2}{c}{10} \\ \cmidrule(l){3-8} 
\multicolumn{2}{c}{Communication round} & \multicolumn{2}{c}{20} & \multicolumn{2}{c}{20} & \multicolumn{2}{c}{20} \\ \cmidrule(l){3-8} 
\multicolumn{2}{c}{Batch size} & \multicolumn{2}{c}{256} & \multicolumn{2}{c}{256} & \multicolumn{2}{c}{64}  \\ \cmidrule(l){3-8} 
  \multicolumn{2}{c}{Epoch} & \multicolumn{2}{c}{1}& \multicolumn{2}{c}{1}& \multicolumn{2}{c}{1} \\ \cmidrule(l){3-8} 
  \multicolumn{2}{c}{Optimizer} & \multicolumn{2}{c}{ADAM}& \multicolumn{2}{c}{ADAM}& \multicolumn{2}{c}{ADAM} \\ \cmidrule(l){3-8} 
  \multicolumn{2}{c}{Learning rate} & \multicolumn{2}{c}{0.002} & \multicolumn{2}{c}{0.002}& \multicolumn{2}{c}{0.001} \\ \cmidrule(l){3-8} 
 \multicolumn{2}{c}{Scheduler} & \multicolumn{2}{c}{N/A}& \multicolumn{2}{c}{N/A} & \multicolumn{2}{c}{MultistepLR} \\ \cmidrule(l){3-8} 
 \multicolumn{2}{c}{Weight decay}& \multicolumn{2}{c}{N/A}& \multicolumn{2}{c}{N/A} &
\multicolumn{2}{c}{0.1}  \\ \bottomrule
\end{tabular}
\end{table}

\subsection{More ablation study}
    Continuing with Subsection 6.2, we present ablation experiments regarding the number of clients and regularization function on the COMPAS dataset.
    
\begin{figure}
        \centering
        \includegraphics[width=0.5\columnwidth]{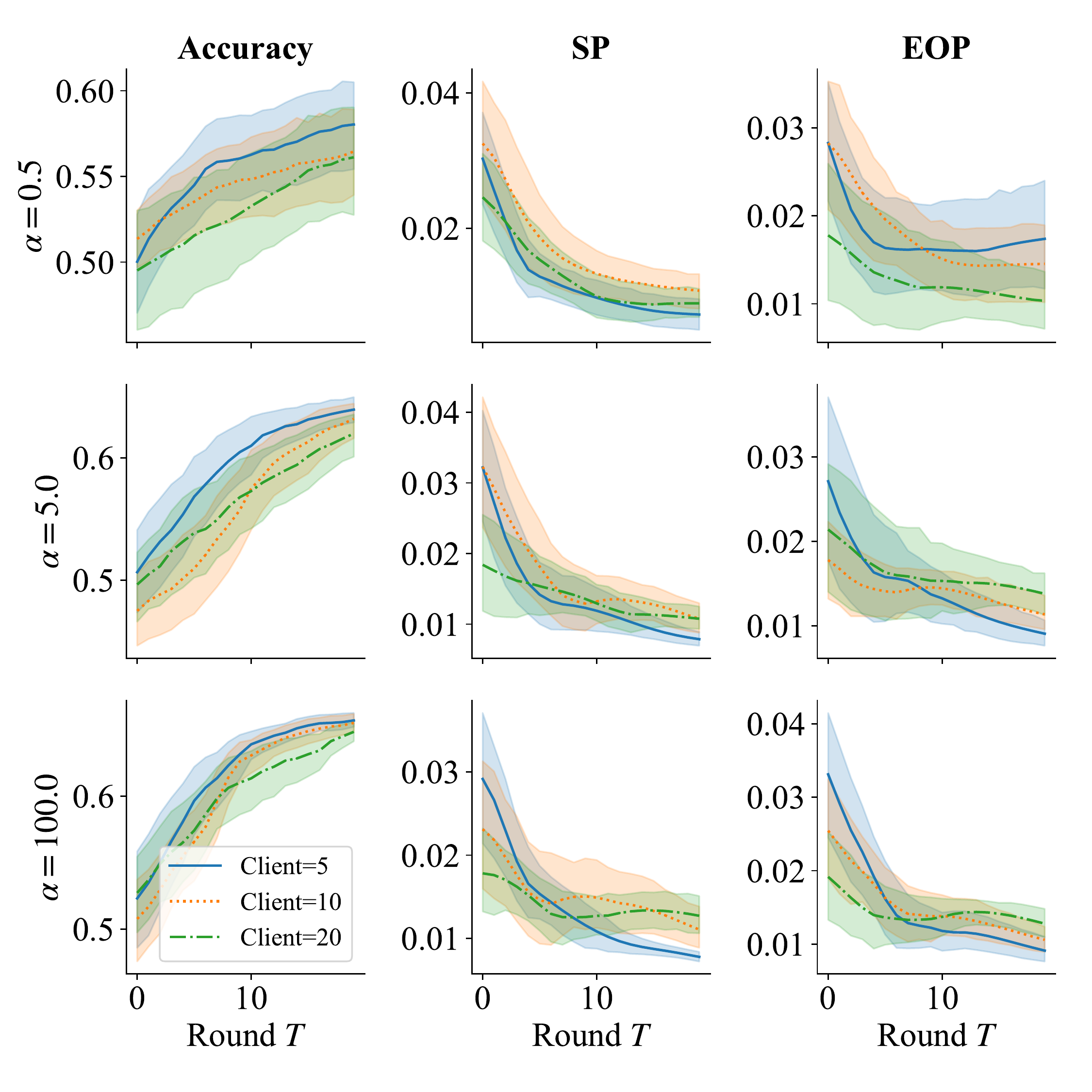}
        \caption{The accuracy and bias of $\mfg$ for different numbers of clients $K$.}
        \label{fig:compas_clients}
    \end{figure}

    \textbf{Number of clients.} $K=\{5, 10, 20\}$ clients are considered, with result in \Autoref{fig:compas_clients}. Overall, the number of clients has little influence on both accuracy and fairness.

    \textbf{Regularization function} The experiments in Section~\ref{sec:exp} use $J(x)=x$, thus the objective function involves the fairness metric, which often contains absolute values. Thus, the optimization may be unstable since absolute functions are non-smooth. To avoid this issue, we propose to use $J(x)=x^2$, thus the penalty term becomes
    \begin{align*}
        J(F(f(\cdot ;\theta); \hat{\mathcal{D}})) = \biggl(\sum_k w_kF_k \biggr)^2,
    \end{align*}
    which is smooth as long as $F_k$'s are smooth. We call $J(x)=x$ as $\ell_1$ penalty and $J(x)=x^2$ as $\ell_2$ penalty, and compare the performance of $\mfg$. 
    The results are reported in \Autoref{tab:reg}. We find that there is no statistically significant difference in both fairness and accuracy. However, the training process of the $\ell_2$ penalty is much more stable than $\ell_1$, and we will recommend using the $\ell_2$ penalty in general.

        \begin{table}[tb]\scriptsize
        \centering
        \caption{The accuracy and bias of $\mfg$ for different regularization functions $J(x)$.}
            \label{tab:reg}
\begin{tabular}{llllll}
\toprule
  \multirow{2}{*}{$\alpha$}    & \multirow{2}{*}{$J(x)$}  & \multicolumn{2}{c}{SP} & \multicolumn{2}{c}{EOP} \\ \cmidrule(l){3-4} \cmidrule(l){5-6}
     &  &           Accuracy &         Bias &           Accuracy &         Bias \\
\midrule
\multirow{2}{*}{$0.5$} 
      & $\ell_1$ &  57.78 (5.53) &  0.26 (0.18) &  56.85 (5.65) &  0.47 (0.64) \\
      & $\ell_2$ &  55.65 (5.87) &  0.47 (0.39) &  56.78 (3.72) &  1.03 (0.57) \\
      \midrule
\multirow{2}{*}{$5$}  
      & $\ell_1$ &  62.27 (4.28) &  0.29 (0.26) &  62.43 (3.38) &  0.32 (0.37) \\
      & $\ell_2$ &  63.04 (2.99) &  0.55 (0.35) &  63.54 (2.73) &  0.51 (0.19)
 \\
      \midrule
\multirow{2}{*}{$100$} 
      & $\ell_1$ &  65.0 (1.44) &  0.35 (0.37) &  65.03 (1.26) &  0.39 (0.43) \\
      & $\ell_2$ &   65.03 (1.28) &  0.72 (0.53) &  64.94 (2.14) &   0.5 (0.26) \\
\bottomrule
\end{tabular}
\end{table}


\subsection{Pure clients}
When clients are purely from one group, local fairness is not well-defined thus locally fair training is not applicable in this situation. Our proposed algorithm is thus preferred in this scenario. We have also conducted additional experiments on the COMPAS dataset to corroborate our algorithm's effectiveness. The results are summarized in \Autoref{tab:pure}. From the results, the proposed algorithm `FedGFT' still mitigates the bias compared to FedAvg, though the accuracy-fairness trade-off is worse than the situation where the clients have data from both groups.

\begin{table}[tb]\small
    \caption{The average accuracy and bias (standard error in parentheses) on the COMPAS dataset under two fairness metrics. Pure group represents the situation where clients are purely from one group; mixed group represents the case where clients have data from both groups; and 
$\lambda$ is the penalty parameter.}
    \label{tab:pure}
    \centering
    \begin{tabular}{llll}
    \toprule
        Method & Acc & SP & EOP \\ \midrule
        FedAvg (Mixed group) &	65.69 (1.76)&	8.04 (1.67)	&	6.71 (1.84) \\ \midrule
FedGFT (Mixed group)&	65.0 (1.44)&	0.35 (0.37)	&	0.39 (0.43) \\ \midrule
FedGFT (Pure group, 
$\lambda=10$)&	62.53 (5.05)&	2.79 (1.47)	&	2.22 (0.9) \\ \midrule
FedGFT (Pure group, 
$\lambda=20$
)	&61.09 (5.02)&	1.83 (1.03)	&	1.56 (0.87) \\ \midrule
FedGFT (Pure group, 
$\lambda=100$
)	&53.0 (6.51)&	0.85 (0.61)&	0.49 (0.26) \\
\bottomrule
    \end{tabular}
\end{table}

\end{document}